\documentclass[sigconf, nonacm]{acmart}

\usepackage{xcolor, color, colortbl}
\usepackage[utf8]{inputenc}
\usepackage{makecell}
\usepackage{amsmath}
\usepackage{bm}
\usepackage{graphicx}
\usepackage{booktabs} 
\usepackage{xspace}
\usepackage{amsmath,bm,bbm}
\usepackage{amsfonts} 
\usepackage{paralist}
\usepackage{mathtools}
\usepackage{tikz}
\usetikzlibrary{calc}
\usepackage{pifont}
\usepackage{subcaption,booktabs}
\usepackage{perpage} 
\MakePerPage{footnote} 
\usepackage[bottom]{footmisc}
\usepackage{enumitem}
\usepackage{makecell}
\usepackage{multirow}
\usepackage{tabularx}
\usepackage{multicol}

\usepackage{xcolor}

\usepackage{listings}

\definecolor{dkred}{rgb}{0.5,0,0}
\definecolor{dkgreen}{rgb}{0,0.6,0}
\definecolor{gray}{rgb}{0.5,0.5,0.5}
\definecolor{mauve}{rgb}{0.58,0,0.82}
\newcommand{\system}{{\sc TOD}\xspace} 

\newcommand{\bX}{\mathbf{X}}

\newcommand{\bD}{\mathbf{D}}
\newcommand{\bI}{\mathbf{I}}

\newcommand{\norm}[1]{\left\lVert#1\right\rVert}

\newtheorem{theorem}{Theorem}

\newcommand{\cbit}{\begin{compactitem}}
	\newcommand{\ceit}{\end{compactitem}}
\newcommand{\cben}{\begin{compactenum}}
	\newcommand{\ceen}{\end{compactenum}}
	
\newcommand\crule[3][black]{\textcolor{#1}{\rule{#2}{#3}}}

\definecolor{aliceblue}{rgb}{0.867, 0.917, 0.964}
\definecolor{aliceyellow}{rgb}{0.999, 0.945, 0.796}
\definecolor{alicegray}{rgb}{0.844, 0.867, 0.898}

\tikzstyle{every number}=[draw=black] 

\newcommand{\cmark}{\ding{51}}%
\newcommand{\xmark}{\ding{55}}%

\usepackage{soul}
\newcommand{\rv}[1]{\textcolor{black}{#1}}

\newcommand\vldbdoi{XX.XX/XXX.XX}
\newcommand\vldbpages{XXX-XXX}
\newcommand\vldbvolume{16}
\newcommand\vldbissue{1}
\newcommand\vldbyear{2022}
\newcommand\vldbauthors{\authors}
\newcommand\vldbtitle{\shorttitle} 
\newcommand\vldbavailabilityurl{URL_TO_YOUR_ARTIFACTS}
\newcommand\vldbpagestyle{plain}
\newcommand{\MyPara}[1]{\noindent\textbf{\textit{#1}}~~}

\begin{document}
\title{\system: GPU-accelerated Outlier Detection via Tensor Operations}

\author{Yue Zhao}
\affiliation{%
  \institution{Carnegie Mellon University}
  \city{Pittsburgh}
  \state{PA}
}
\email{zhaoy@cmu.edu}

\author{George H. Chen}
\affiliation{%
  \institution{Carnegie Mellon University}
  \city{Pittsburgh}
  \state{PA}
}
\email{georgechen@cmu.edu}

\author{Zhihao Jia}
\affiliation{%
  \institution{Carnegie Mellon University}
  \city{Pittsburgh}
  \state{PA}
}
\email{zhihao@cmu.edu}

\begin{abstract}
\rv{Outlier detection (OD) is a key learning task for finding rare and deviant data samples, with many time-critical applications such as fraud detection and intrusion detection.}
In this work, we propose \system, the first tensor-based system for efficient and scalable outlier detection
on distributed multi-GPU machines.
A key idea behind \system is decomposing complex OD applications into a small collection of basic tensor algebra operators. This decomposition enables \system to accelerate OD computations by leveraging recent advances in deep learning infrastructure in both hardware and software.
Moreover, to deploy memory-intensive OD applications on modern GPUs with limited on-device memory, we introduce two key techniques.
First, {\em provable quantization} speeds up OD computations and reduces its memory footprint by automatically performing specific floating-point operations in lower precision while provably guaranteeing no accuracy loss.
Second, to exploit the aggregated compute resources and memory capacity of multiple GPUs, we introduce {\em automatic batching}, which decomposes OD computations into small batches for both sequential execution on a single GPU and parallel execution on multiple GPUs.

\system supports a diverse set of OD algorithms.
\rv{Extensive evaluation on 11 real and 3 synthetic OD datasets shows that \system is on average 10.9$\times$ faster than the leading CPU-based OD system PyOD (with a maximum speedup of 38.9$\times$), and can handle much larger datasets than existing GPU-based OD systems.}
In addition, \system allows easy integration of new OD operators, enabling fast prototyping of emerging and yet-to-discovered OD algorithms. 
\end{abstract}

\maketitle

\pagestyle{\vldbpagestyle}
\begingroup\small\noindent\raggedright\textbf{PVLDB Reference Format:}\\
\vldbauthors. \vldbtitle. PVLDB, \vldbvolume(\vldbissue): \vldbpages, \vldbyear.\\
\href{https://doi.org/\vldbdoi}{doi:\vldbdoi}
\endgroup
\begingroup
\renewcommand\thefootnote{}\footnote{\noindent
This work is licensed under the Creative Commons BY-NC-ND 4.0 International License. Visit \url{https://creativecommons.org/licenses/by-nc-nd/4.0/} to view a copy of this license. For any use beyond those covered by this license, obtain permission by emailing \href{mailto:info@vldb.org}{info@vldb.org}. Copyright is held by the owner/author(s). Publication rights licensed to the VLDB Endowment. \\
\raggedright Proceedings of the VLDB Endowment, Vol. \vldbvolume, No. \vldbissue\ %
ISSN 2150-8097. \\
\href{https://doi.org/\vldbdoi}{doi:\vldbdoi} \\
}\addtocounter{footnote}{-1}\endgroup

\ifdefempty{\vldbavailabilityurl}{}{
\vspace{.3cm}
\begingroup\small\noindent\raggedright\textbf{PVLDB Artifact Availability:}\\
The source code, data, and other artifacts have been made available at \url{https://github.com/yzhao062/pytod}.
\endgroup
}

\section{Introduction}
\label{sec:intro}
\rv{Outlier detection (OD) is a crucial machine learning task for identifying data points deviating from a general distribution \cite{aggarwal2015outlier,yu2016survey,liu2021fairness}. OD has numerous real-world applications, including 
anti-money laundering \cite{DBLP:conf/bigdataconf/Lee0WLATF20}, rare disease detection \cite{pang2021deep}, rumor detection \cite{tam2019anomaly}, and network intrusion detection \cite{lazarevic2003comparative}.
OD algorithms have been serving a critical role in large cloud services for monitoring server abnormality at Microsoft \cite{ren2019time} and Amazon \cite{ayed2020anomaly}, as well as for fraud detection at eBay \cite{abdulaal2021practical} and Alibaba \cite{liu2021intention}}. 

\noindent\rv{\MyPara{Scalability challenges of OD}. Numerous OD algorithms have been proposed recently to detect outliers for different types of data (e.g., tabular data \cite{aggarwal2015outlier,kingsbury2020elle,li2022ecod,zhao2019lscp}, time series \cite{boniol2020series2graph,lai2021revisiting,boniol2021sand,cao2019efficient}, graphs \cite{aggarwal2011outlier,boniol2020graphan}).
Although there is no shortage of detection algorithms, 
OD applications face challenges in \textit{scaling to large datasets}, both in terms of execution time and memory consumption, which prevents OD algorithms from being deployed in data-intensive or time-critical tasks such as real-time credit card fraud detection.}
To address these challenges, recent work focuses on both developing distributed OD algorithms on CPUs \cite{lozano2005parallel,otey2006fast,angiulli2010distributed,bhaduri2011algorithms,oku2014parallel,yan2017distributed,toliopoulos2020proud,zhao2021suod,zhang2022sparx} and accelerating certain OD algorithms on GPUs \cite{DBLP:journals/tpds/AngiulliBLS16,DBLP:journals/sigkdd/LealG18}. However, existing GPU-based OD solutions only target specific (families of) OD algorithms and cannot support {\em generic} OD computations. 
For instance, Angiulli et al. \cite{DBLP:journals/tpds/AngiulliBLS16} showcases an example of using GPUs for distance-based algorithms, while how to handle linear and probabilistic OD algorithms remains unclear under the proposed solution.

\MyPara{Advances in deep neural networks}. On the other hand, deep neural networks (DNNs) have revolutionized computer vision, natural language processing, and various other fields~\cite{lecun2015deep,goodfellow2016deep,Yang2022DiffusionMA} over the last decade.
This success is largely due to the recent development of DNN systems (e.g., TensorFlow~\cite{abadi2016tensorflow} and PyTorch~\cite{paszke2019pytorch}). These systems achieve fast tensor algebra computations (e.g., matrix multiplication, convolution, etc.) on modern hardware accelerators (e.g., GPUs and TPUs) and use efficient parallelization strategies (e.g., data, model, and pipeline parallelism~\cite{abadi2016tensorflow, FlexFlow, PipeDream}) to aggregate the compute resources across multiple accelerators, enabling efficient
and scalable DNN computations.

This paper explores a new approach to building GPU-accelerated OD systems. Instead of following the methodology used in existing GPU-based OD frameworks (i.e., providing efficient GPU implementations tailored to specific OD applications), we ask: {\em can we leverage the compilation and optimization techniques in DNN systems to minimize the time and memory consumption of a wide range of common OD computations?}

\begin{figure}[!t] 
\centering
  \includegraphics[width=0.48\textwidth]{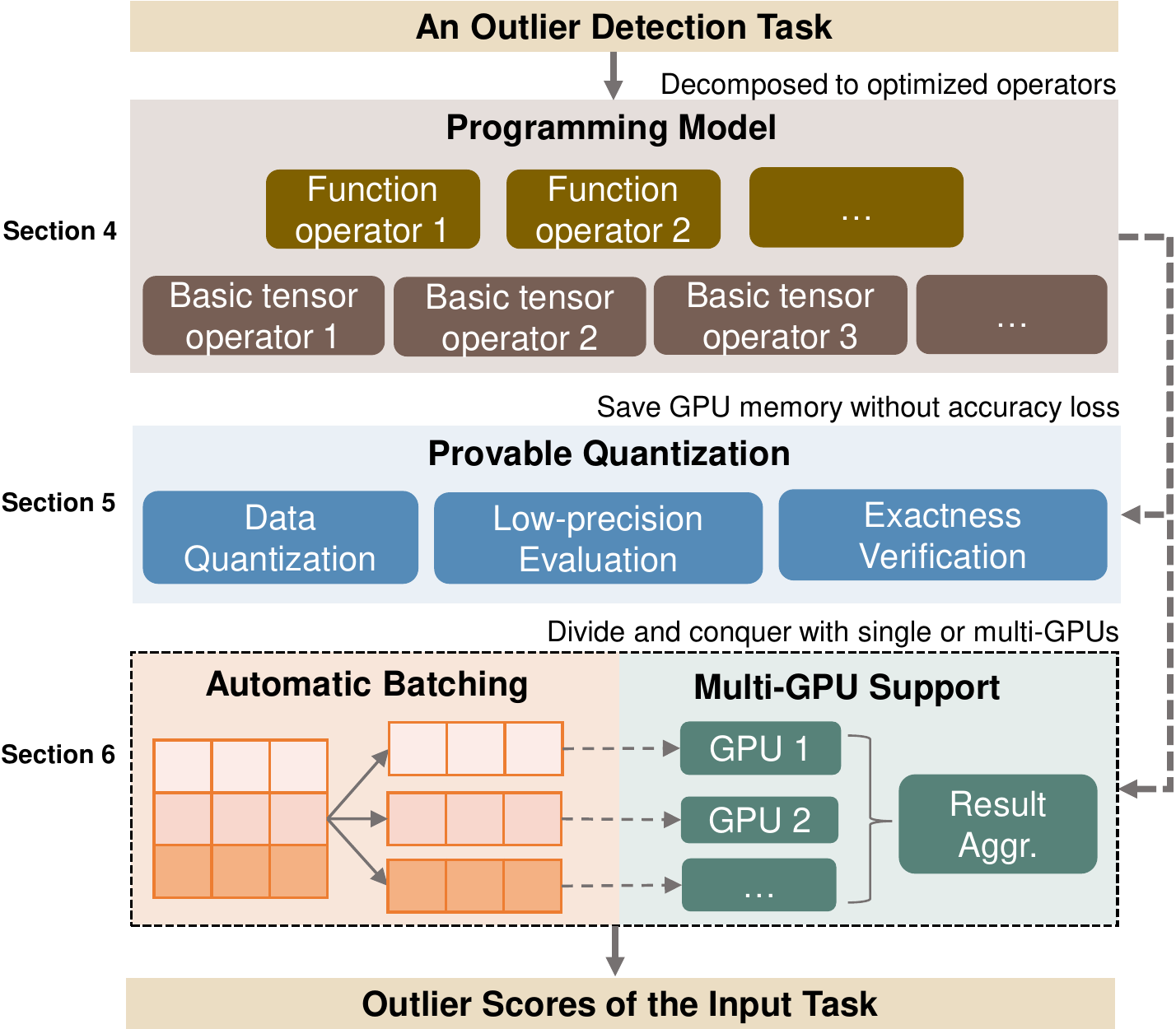}
  \caption{System overview. \system abstracts OD algorithms into small building blocks via the \textit{programming model} for deep optimization, with a set of techniques including \textit{provable quantization}, \textit{automatic batching}, and \textit{multi-GPU support}.}
 \label{fig:overview}
\end{figure}

\subsection{Our Approach}
In this paper, we present \system, a \underline{t}ensor-based \underline{o}utlier \underline{d}etection system that abstracts OD applications into tensor operations for efficient GPU acceleration.
\system leverages both the software and hardware optimizations in modern deep learning frameworks to enable efficient
and scalable OD computations on distributed multi-GPU clusters.
To the best of our knowledge, \system is the \textit{first} GPU-based system for {\em generic} OD applications.
Fig.~\ref{fig:overview} shows an overview of \system.
Building a tensor-based OD system requires addressing three major obstacles.

\MyPara{Representing OD computations using tensor operations.}
Unlike DNN models, which are represented as a pipeline of tensor algebra operators (e.g., matrix multiplication), many OD applications involve a diverse collection of operators that have traditionally not been implemented in terms of tensor operations, such as proximity-based algorithms, statistical approaches, and linear methods (see an overview of OD applications in \S \ref{subsec:od_algorithms}). Implementing OD applications one at a time and accounting for software and hardware optimizations is labor-intensive.
To address this obstacle, \system introduces a new programming model for OD that decomposes a broad set of OD applications into a small collection of {\em basic tensor operators} and {\em functional operators}, which significantly reduces the implementation and optimization effort and opens the possibility of easily supporting new OD algorithms.

\MyPara{Quantizing OD computations.}
{\em Quantization} is commonly used in existing DNN frameworks to reduce time and memory consumption of DNN computations by executing intermediate operations of DNN models in a low-precision floating-point representation.
Quantization in general does not preserve the end-to-end equivalence of a DNN model and may introduce potential accuracy losses.
To apply quantization, the current practice is fine-tuning a quantized DNN model on the training dataset in a {\em supervised} fashion and assessing the accuracy loss; however, this approach is not directly applicable to OD, since most OD algorithms are {\em unsupervised} and thus lack a direct way for measuring accuracy.
To address this challenge, \system introduces a novel quantization technique called {\em provable quantization}, which leverages the numerical insensitivity of some OD algorithms (e.g., $k$-nearest-neighbors may return identical results for some samples when computing with different floating-point precisions) and {\em automatically} performs specific OD operators in lower precision. In contrast to prior quantization techniques that also use lower precision calculations at the expense of accuracy, our proposed provable quantization technique provably guarantees {\em no} accuracy loss.

\MyPara{Enabling scalable OD computations.}
Existing deep learning frameworks cannot directly support large-scale OD applications, since modern deep learning systems are designed to iteratively process a small {\em batch} of training samples even though the entire training dataset can be large.
For example, to train ResNet-50~\cite{resnet} on the ImageNet dataset~\cite{imagenet}, the DNN systems only handle small mini-batches (e.g., 256 samples) in each training iteration, while the dataset contains more than 14 million samples.
However, many OD applications involve operating on {\em all} samples, such as computing distances between all sample pairs. Executing such an application on a single GPU would typically run out of memory because GPUs nowadays have limited memory capacities (e.g., compared to those of CPU RAM).
To overcome the difficulty of executing OD applications in iterations and the resource limits of a single GPU, \system uses an {\em automatic batching} mechanism to execute memory-intensive OD operators in small batches, which are distributed across {\em multiple} GPUs in parallel in a pipeline fashion.
The automatic batching and multi-GPU support allow \system to scale to datasets as large as those commonly encountered in deep learning.

We compare \system against existing CPU- and GPU-based OD systems on both real-world and synthetic datasets. \system is on average 10.9 times faster than PyOD, a state-of-the-art comprehensive CPU-based OD system~\citep{DBLP:journals/jmlr/ZhaoNL19}, and can process a million samples within an hour while PyOD cannot. Compared to existing GPU-based OD systems, \system can handle much larger datasets, while the GPU baselines run out of GPU memory.
Our evaluation further shows that provable quantization, automatic batching, and multi-GPU support are all critical for efficient and scalable OD computations.

In summary, this paper makes the following contributions:
\begin{itemize}
[leftmargin=*]
\item We propose \system, the first tensor-based system for generic outlier detection, enabling efficient
and scalable OD computations on distributed multi-GPU clusters.
\item \system uses a new programming model that abstracts complex OD applications into a small collection of basic tensor operators for efficient GPU acceleration.
\item We introduce provable quantization that accelerates unsupervised OD computations by performing specific floating-point operators in lower precision while provably guaranteeing no accuracy loss.

\end{itemize}

\MyPara{Extensibility and integration.}
\system is open-sourced\footnote{Open-sourced Library: \url{https://github.com/yzhao062/pytod}} (see Appx. \S \ref{sec:api} for API demonstration), which enables easy development of new OD algorithms by leveraging highly optimized tensor operators or including new operators (see examples in \S \ref{subsec:implement}). 
This extensibility of \system yields a large number of yet-to-be-discovered OD methods. Thus, we believe that \system also provides a platform that enables rapid research and development of novel OD methods.

\section{Background and Related Work}
\label{sec:related_work}
We provide background on existing OD algorithms for tabular data and which ones are accelerated by \system (\S \ref{subsec:od_algorithms}), on the DNN infrastructure that \system builds off of to do acceleration (\S \ref{subsec:dl-infrastructure}), and on existing comprehensive OD systems (\S \ref{subsec:od-sys}), including the current state-of-the-art PyOD \citep{DBLP:journals/jmlr/ZhaoNL19}. 
In \S \ref{subsec:other_sys}, we review additional systems,  algorithms, and applications for other data formats in addition to tabular data (e.g., time series and graphs). Note that \system primarily focuses on OD in tabular data due to its popularity \cite{aggarwal2015outlier}; \system can be extended to support OD in other data types with modifications.

\begin{table}[!t] 
\centering
	\caption{Key OD algorithms for tabular data and their time and space complexity with a \textit{brute-force} implementation (additional optimization is possible but not considered here), where $n$ is the number of samples, and $d$ is the number of dimensions. Note that ensemble-based methods' complexities depend on the underlying base estimators. Algorithms that can be accelerated in \system are marked with \cmark.} 
\begin{tabular}{ll|ccc}
    \toprule
    \textbf{Category} & \textbf{Algorithm} & \makecell{\textbf{Time} \\ \textbf{Compl.}} & 
    \makecell{\textbf{Space} \\ \textbf{Compl.}} & \makecell{\textbf{Optimized} \\ \textbf{in \system}} \\
    \midrule
    Proximity         & $k$NNOD              & $O(n^2)$  & $O(n^2)$  & \cmark                  \\
    Proximity         & COF                & $O(n^3)$   & $O(n^2)$  & \cmark                  \\
    Proximity         & LOF                & $O(n^2)$   & $O(n^2)$   & \cmark                  \\
    Proximity         & LOCI               & $O(n^2)$   & $O(n^2)$  & \cmark                  \\
    Statistical       & KDE                & $O(n^3)$ & $O(n^2)$                      & \xmark                \\
    Statistical       & HBOS               & $O(nd)$  & $O(nd)$                     & \cmark                  \\
    Statistical       & COPOD              & $O(nd)$  & $O(nd)$                     & \cmark                  \\
    Statistical       & ECOD              & $O(nd)$  & $O(nd)$                     & \cmark                  \\
    Ensemble          & LODA               & N/A     & N/A                    & \cmark                 \\
    Ensemble          & FB                 & N/A    & N/A                    &
    \cmark                 \\
    Ensemble          & iForest            & N/A    & N/A                  & \xmark \\
    Ensemble          & LSCP               & N/A     & N/A                   &  \cmark                   \\
    Linear            & PCA                & $O(nd)$    & $O(n)$ & \cmark                  \\
    Linear            & OCSVM              & $O(n^3)$   & $O(n^2)$ & \xmark     \\
    \bottomrule
\end{tabular}
	\label{table:algorithms} 
 \vspace{-0.15in}
\end{table}

\subsection{Existing OD Algorithms and Scalability}
\label{subsec:od_algorithms}

Outlier detection (also called anomaly detection) is a key machine learning task that aims to find data points that deviate from a general distribution \cite{aggarwal2015outlier,yu2016survey,liu2021fairness}. 
As shown in Table \ref{table:algorithms}, non-deep-learning OD algorithms may be grouped into four categories (see the book \cite{aggarwal2015outlier} by Aggarwal for more details on algorithms):
 (\textit{i}) proximity-based algorithms that rely on measuring sample similarity including $k$NN \cite{Angiulli2002fast}, ABOD \cite{kriegel2008angle}, 
 COF \cite{DBLP:conf/pakdd/TangCFC02}, 
 LOF \cite{Breunig2000}, and LOCI \cite{DBLP:conf/icde/PapadimitriouKGF03}; (\textit{ii}) statistical approaches including 
 KDE \cite{DBLP:conf/sdm/SchubertZK14}, 
 HBOS \cite{goldstein2012histogram}, COPOD \cite{DBLP:conf/icdm/LiZBIH20}, and ECOD \cite{li2022ecod}; 
 (\textit{iii}) ensemble-based methods that build a collection of detectors for aggregation like iForest \cite{DBLP:conf/icdm/LiuTZ08}, LODA \cite{pevny2016loda}, 
 and LCSP \cite{zhao2019lscp}; and (\textit{iv}) linear models such as PCA \cite{shyu2003novel}.

Importantly, many OD algorithms suffer from scalability issues \cite{orair2010distance,zhao2021suod}. For example, Table \ref{table:algorithms} shows that various proximity-based OD algorithms have at least $O(n^2)$ time and space complexities---they all require estimating and storing pairwise distances among all $n$ samples. 
The high time complexities of many OD algorithms limit their applicability in real-world applications that require either real-time responses (e.g., fraud detection \cite{DBLP:conf/sigir/LiuDYDP20}) or the concurrent processing of millions of samples \cite{zhong2020financial}. 
As shown in the table, 
\system supports nearly all the OD algorithms mentioned.

\subsection{DNN Infrastructure and Acceleration} 
\label{subsec:dl-infrastructure}

Deep neural networks have dramatically improved the accuracy of artificial intelligence systems across numerous fields \cite{goodfellow2016deep,fu2021probabilistic,pang2021deep}. Its success is fueled by recent advances in both hardware and software~\cite{lecun20191}. Specifically, DNN systems depend on tensor operations that can often be parallelized and executed in small batches. These operations are well-suited for GPUs, especially as a single GPU nowadays often has many more cores than a single CPU; while GPU cores are not as general purpose as CPU cores, they suffice in executing the tensor operations of deep learning. Moreover, the maturity of DNN programming interfaces such as PyTorch \cite{paszke2019pytorch} and TensorFlow \cite{abadi2016tensorflow} makes developing machine learning models easy with a wide range of GPUs. Multiple works attempt to leverage DNN systems for accelerating training data science and ML tasks, including Hummingbird \cite{DBLP:conf/osdi/NakandalaSYKCWI20}, Tensors \cite{koutsoukos2021tensors}, and AC-DBSCAN \cite{ji2021accelerating}.

Differently, \system for the first time, extends the acceleration usage of DNN systems to OD algorithms.
We build \system using the DNN ecosystem, taking advantage of its established hardware acceleration and software accessibility. This design choice also opens the opportunity for unifying classical OD algorithms (see \S \ref{subsec:od_algorithms}) and DNN-based OD algorithms on the same platform---this emerging direction has gained increasing attention in OD research \cite{ruff2019deep}.

\subsection{Outlier Detection Systems}
\label{subsec:od-sys}

\MyPara{CPU-based systems}. Over the years, comprehensive OD systems on CPUs that cover a diverse group of algorithms have been developed in different programming languages, including ELKI Data Mining \citep{Achtert2010visual} and
RapidMiner \citep{Hofmann2013rapidminer} in Java, and PyOD \citep{DBLP:journals/jmlr/ZhaoNL19} in Python. Among these, PyOD is the state-of-the-art (SOTA) with deep optimization including just-in-time compilation and parallelization. It is widely used in both academia and industry, with hundreds of citations \cite{pyodcite} and millions of downloads per year \cite{pyodpepy}. Recently, the PyOD team proposed an acceleration system called SUOD to further speed up the training and prediction in PyOD with a large collection of heterogeneous OD models \cite{zhao2021suod}. Specifically, SUOD uses algorithmic approximation and efficient parallelization to reduce the computational cost and therefore runtime. There are other distributed/parallel systems designed for specific (family of) OD algorithms with non-GPU nodes (e.g., CPUs): (\textit{i}) Parallel Bay, Parallel LOF, DLOF for local OD algorithms \cite{lozano2005parallel, oku2014parallel,yan2017distributed}, (\textit{ii}) DOoR for distance-based OD \cite{bhaduri2011algorithms}, (\textit{iii}) distributed OD for mixed-attributed data \cite{otey2006fast} (\textit{iv}) PROUD for stream data \cite{toliopoulos2020proud} and (\textit{v}) Sparx for Apache Spark \cite{zhang2022sparx}. These distributed non-GPU systems do not constitute as baselines as \system is a comprehensive system covering different types OD algorithms, while the specialized systems only cover specific algorithms. Thus, we consider the SOTA comprehensive system, PyOD, as the primary baseline (see exp. results in \S \ref{subsec:end_2_end}).

\begin{figure*}[!t] 
\centering
  \includegraphics[width=0.91\textwidth]{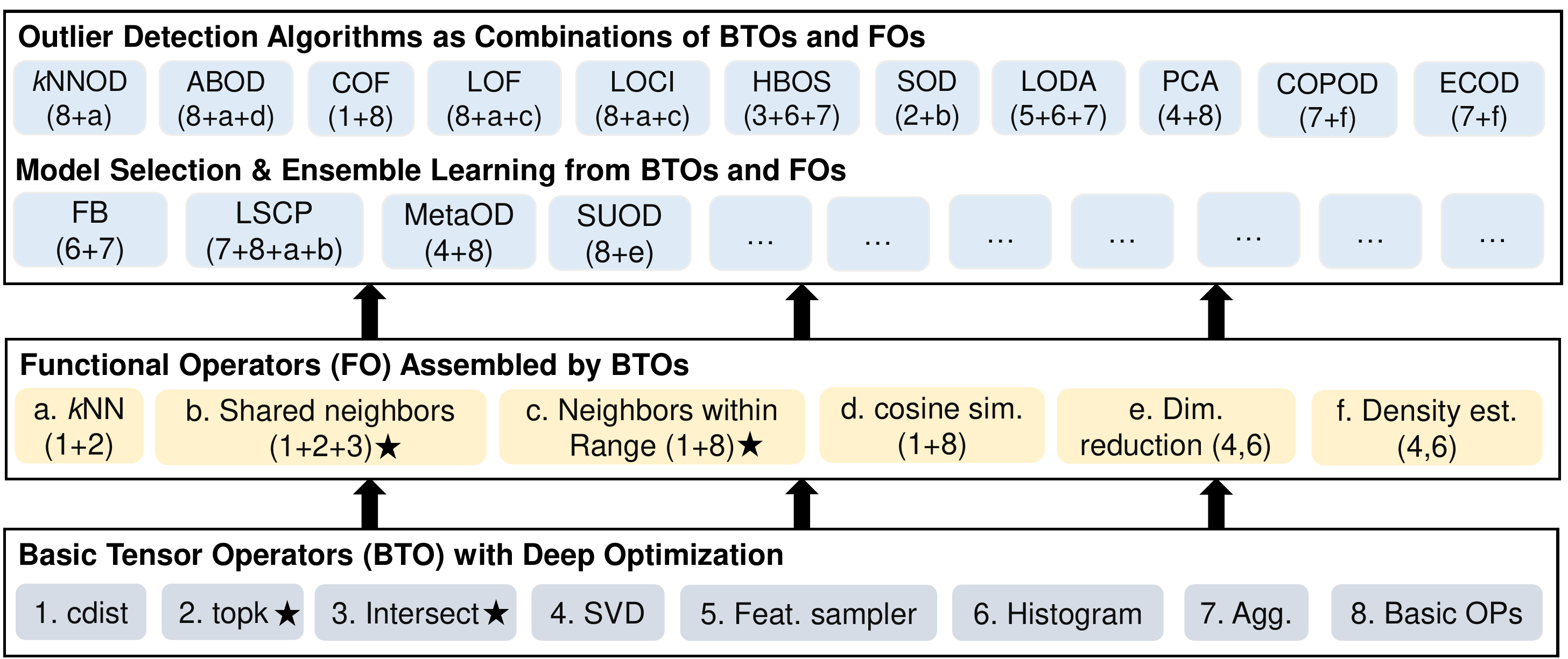}
  \caption{With algorithmic abstraction, more than 20 OD algorithms (denoted by \crule[aliceblue]{0.2cm}{0.2cm}) are abstracted into eight basic tensor operators (in \crule[alicegray]{0.2cm}{0.2cm}) and six functional operators (in \crule[aliceyellow]{0.2cm}{0.2cm}). This abstraction reduces the implementation and optimization effort, and opens the possibility of including new algorithms.
  \rv{All operators are executed on GPUs using automatic batching (see \S \ref{sec:batching})}, and operators marked with $\bigstar$ are further accelerated using provable quantization (see \S \ref{sec:quantization}).}
\label{fig:operators}
\end{figure*}

\smallskip
\noindent \MyPara{GPU-based systems}. There are efforts to use GPUs for fast OD calculations for LOF \cite{alshawabkeh2010accelerating}, distance-based methods \cite{DBLP:journals/tpds/AngiulliBLS16}, KDE \cite{azmandian2012gpu}, and data stream  \cite{DBLP:journals/sigkdd/LealG18}. These approaches rely on exploring the characteristics of a specific OD algorithm for GPU acceleration. This limits their generalization to a wide collection of OD algorithms. Furthermore, none has direct multi-GPU support, leading to limited scalability. To the best of our knowledge, there is no existing comprehensive GPU-based OD system that covers a diverse group of algorithms. 
Thus, we use direct GPU implementations of OD algorithms and selected works above as GPU baselines when appropriate (see details in \S \ref{subsec:data_base_metrics} and Table \ref{table:gpu}).

\subsection{Systems for Other Data Types and Scenarios}
\label{subsec:other_sys}
Over the years, various algorithms and systems have been developed for OD with different types of data other than tabular, including time-series/sequence (TOP \cite{cao2019efficient}, NETS \cite{yoon2019nets}, GraphAn \cite{boniol2020graphan}, CPOD \cite{tran2020real}, Series2graph \cite{boniol2020series2graph}, SAND \cite{boniol2021sand}, etc.), and graph (e.g., Elle \cite{kingsbury2020elle}, PyGOD \cite{liu2022pygod,liu2022benchmarking}, etc.). Also, different input data are also assumed in streaming and feature-evolving fashions (e.g., xStream \cite{manzoor2018xstream}, etc.). 
Meanwhile, emphasis has also been given to building systems for explaining outliers (VSOutlier \cite{cao2014interactive} and Exathlon \cite{jacob2021demonstration}, etc.) and outlier repairing (IMR \cite{zhang2017time}, etc.), other than detecting them. In this paper, we focus on the detection task in the most prevalent setting (i.e., static tabular data) \cite{aggarwal2015outlier}, while future works may extend to other data types and scenarios.

\section{Overview}
\label{sec:system_overview}
\subsection{Definition and Problem Formulation}
A comprehensive OD system implements a collection of OD models $\mathcal{M}=\{M_1,...,M_m\}$ such that given a user-specified OD model $M \in \mathcal{M}$ and input data $\mathbf{X} \in \mathbbm{R}^{n \times d}$ \textit{without} ground truth labels (rows of $\mathbf{X}$ index data points, while columns index features), the system outputs outlier scores $\mathbf{O} := M(\mathbf{X}) \in \mathbbm{R}^{n}$, which should be roughly deterministic and 
irrespective of the underlying system (higher values in $\mathbf{O}$ correspond to data points more likely to be identified as outliers; threshold on outlier scores can be used to determine which points are outliers). Given a hardware configuration $\mathcal{C}$, e.g., CPU, RAM, and GPU (if available), the system performance can be measured in \textit{efficiency} (both runtime and memory consumption).

\subsection{System Overview}
As a reminder from Section~\ref{sec:intro},
\system is a comprehensive (i.e., covering a diverse group of methods) OD system as outlined in Fig.~\ref{fig:overview} and Table \ref{table:algorithms}. For an outlier detection task, \system decomposes it into a combination of predefined tensor operators via the proposed \textit{programming model} for direct GPU acceleration (\S \ref{sec:model_abstraction}). Notably, \system opportunistically performs {\em provable quantization} on tensor operators to enable faster computation and reduce memory requirements, while provably maintaining model accuracy (\S \ref{sec:quantization}). To overcome the resource limitation of a single GPU, we further introduce \textit{automatic batching} and \textit{multi-GPU} support in \system (\S \ref{sec:batching}).

\section{Programming Model}
\label{sec:model_abstraction}

\MyPara{Motivation.} As a comprehensive system, \system aims to include a diverse collection of OD algorithms, including proximity-based methods, statistical methods, and more (see \S \ref{subsec:od_algorithms}). However, not all algorithms can be directly converted into tensor operations for GPU acceleration.  A key design goal of our programming model is to allow for many OD algorithms to be implemented by piecing together some basic commonly recurring building blocks. In particular, rather than manually implementing many OD algorithms separately, which is a labor-intensive process, we instead define OD algorithms in terms of basic building blocks that each just needs to be implemented once. Moreover, the building blocks can be optimized independently.

\subsection{Algorithmic Abstraction}
\label{subsec:algo_abstract}

The key idea of our programming model is to decompose existing OD algorithms into a set of low-level {\em basic tensor operators} (BTOs), which can directly benefit from GPU acceleration. On top of these BTOs, we introduce higher-level OD operators called {\em functional operators} (FOs) with richer semantics. Consequently, OD algorithms can be constructed as combinations of BTOs and FOs.

Fig.~\ref{fig:operators} shows the hierarchy of \system's programming abstraction in a bottom-up way, with increasing dependency: 8 BTOs are first constructed as the foundation of \system (shown at the bottom in gray), while 6 FOs are then created on top of them (shown in the middle). Finally, OD algorithms and key functions on the top of the figure can be assembled by BTOs and FOs. In other words, \system uses a tree-structured dependency graph, where the BTOs (as leaves of the tree) are fully independent for deep optimization, and all the algorithms depending on these BTOs can be collectively optimized. Clearly, this abstraction process reduces the repetitive implementation and optimization effort, and improves system efficiency and generalization. Additionally, it also facilities fast prototyping and experimentation with new algorithms.

\subsection{Building Complete Algorithms}
\label{subsec:complete_algorithm}

\begin{figure}[t] 
\centering
\subfloat[Abstraction of ABOD]{%
  \includegraphics[clip,width=0.5\columnwidth]{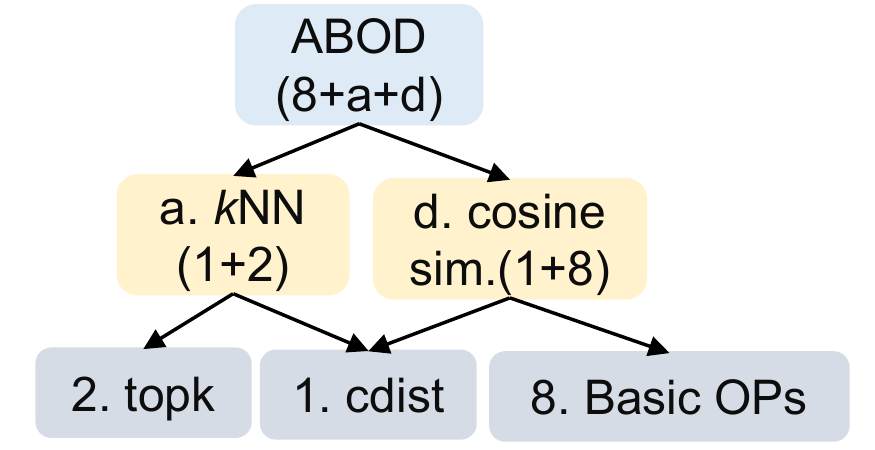}%
}
\subfloat[Abstraction of COPOD]{%
  \includegraphics[clip,width=0.5\columnwidth]{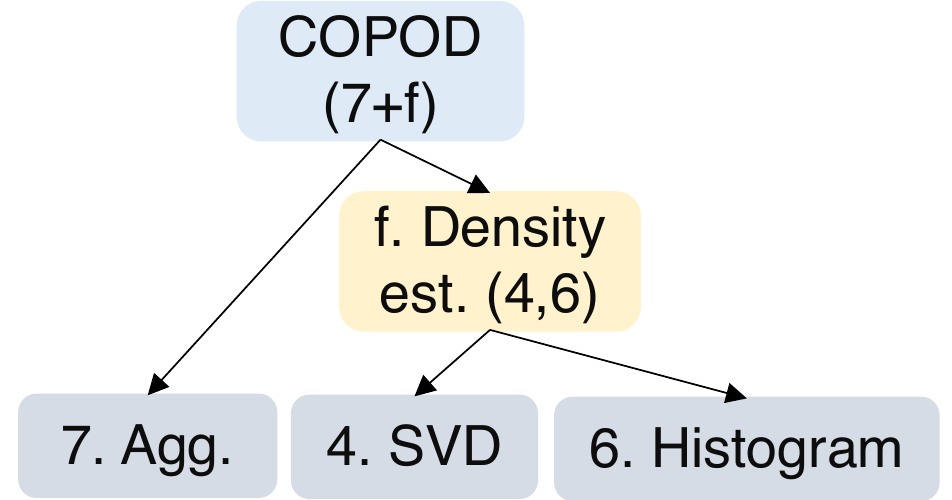}%
}
\caption{Examples of building complex OD algorithms with FO and BTO conveniently.}
\label{fig:abstraction_examples}
\end{figure}

It is easy to build an end-to-end OD application by constructing its {\em computation graph} using FOs and BTOs. For instance, angle-based outlier detection (ABOD) is a classical OD algorithm \cite{kriegel2008angle}, where each sample's outlier score is calculated as the average cosine similarity of its neighbors.
Fig.~\ref{fig:abstraction_examples}(a) highlights an implementation of ABOD that uses an FO called \texttt{$k$NN} to obtain a list of neighbors for each sample and then applies another FO called \texttt{cosine sim.} for calculating cosine similarity. Note that the FOs are also built as combinations of BTOs. For example, \texttt{$k$NN} is implemented as calculating pairwise sample distance using \texttt{cdist} and then identifying the $k$ ``neighbor'' with smallest distances using \texttt{topk}. Additionally, Fig.~\ref{fig:abstraction_examples}(b) shows the abstraction graph of copula-based outlier detection (COPOD) \cite{DBLP:conf/icdm/LiZBIH20}. Other OD algorithms follow the same abstraction protocol as a combination of BTOs and FOs.

\section{Provable Quantization}
\label{sec:quantization}
\MyPara{Motivation.} OD operators mainly depend on floating-point operations, namely $\{+,-,\times, /\}$. For these operations, the main source of the imprecision is rounding \cite{DBLP:journals/pacmpl/Lee0A18}. Of course, rounding errors increase when storing numbers using fewer bits, e.g., 16-bit precision leads to more inaccuracy than 64-bit precision. Many machine learning algorithms therefore use high-precision floating-points when possible to minimize the impact of the rounding errors. However, working with high-precision floating-point numbers can increase computation time and storage costs. This is especially critical for GPU systems with limited memory.

To reduce memory usage and runtime, {\em quantization} has been applied to many machine learning algorithms \cite{blalock2017bolt} and data-driven applications \cite{andre2016cache,wang2020deltapq,wang2020ppq}.
Simply put, it refers to executing an operator (function) with  lower-precision floating representations. If we denote the original function by $r(x)$ and its quantization by $r_q(x)$, the rounding error $\text{Err}(\cdot)$ of quantization is defined as the output difference between $r(x)$ and $r_{q}(x)$, namely $\text{Err}(r_{q}(x)) = r(x)- r_{q}(x)$. Intuitively, quantization can save memory at the cost of accuracy. How to balance the tradeoff between the \textit{memory cost} and \textit{algorithm accuracy} is a key challenge for quantizing in machine learning \cite{DBLP:journals/corr/abs-2102-04503}.
In supervised ML, one may measure the inaccuracy caused by quantization via using ground truth labels to evaluate the performance. However, this is infeasible under unsupervised OD settings, where no ground truth labels is available for evaluation as described in \S \ref{sec:system_overview}. Thus, existing quantization techniques  for supervised ML do not suit the need of unsupervised OD.

In \system, we design a correctness-preserving quantization for (unsupervised) OD applications, termed \textit{provable quantization}. The key idea we use is that depending on the operator used, it is possible to apply quantization to save memory consumption with \emph{no} loss in accuracy. As a motivating example, consider the sign function $r(x)$ that returns ``$+$'' if $x>0$ and returns ``$-$'' otherwise. Clearly, even if we quantize $x$ to have a single bit (that precisely indicates the sign of $x$), we can achieve an exact answer for $r(x)$ that is the same as if instead $x$ had more bits. Similarly, the ranking between two floating-point numbers often only depends on the most significant digits. Building on this simple intuition, we introduce \emph{provable quantization} for a collection of OD operators, where the output and accuracy of the operators remain provably unchanged before and after quantization.

\subsection{$(1+\epsilon)$-property for Rounding Errors}

Provable quantization relies on a standard analysis technique for floating-point numbers called the ``$(1+\epsilon)$-property'' (e.g., \cite{DBLP:journals/pacmpl/Lee0A18}).
Let $\mathbb{F}$ denote the set of 64-bit floating-point numbers. For $x,y\in\mathbb{F}$, we define the floating-point operation ``$\circledast$'' as $x \circledast y \triangleq \text{fl}(x *y)$, where  $* \in \{+, -, \times, /\}$ and $\text{fl}(\cdot)$ refers to the IEEE 754 standard for rounding a real number to a 64-bit floating-point number \cite{kahan1996ieee}. For example, $\oplus$ is floating-point addition and $\otimes$ is floating-point multiplication. 
The standard technique for calculating the rounding errors in floating-point operations is the $(1+\epsilon)$-property \cite{DBLP:journals/pacmpl/Lee0A18}, which is formally defined as follows.
\begin{theorem}[Theorem 3.2 of~\citealt{DBLP:journals/pacmpl/Lee0A18}]
Let $x,y\in\mathbb{F}$, and $*\in\{+,-,\times,/\}$. Suppose that $|x*y|\le\max\mathbb{F}$. Then when we compute $x*y$ in floating-point, there exist multiplicative and additive error terms $\delta\in\mathbb{R}$ and $\delta'\in\mathbb{R}$ respectively such that
\begin{equation}
x\circledast y=(x*y)(1+\delta)+\delta',\quad\text{where }|\delta|\le\epsilon,|\delta'|\le\epsilon'.
\end{equation}
In the above equation, $\epsilon$ and $\epsilon'$ are constants that do not depend on $x$ or $y$. For instance, when working with 64-bit floating-point numbers, $\epsilon=2^{-53}$ and $\epsilon'=2^{-1075}$.
\label{theorem:quantization}
\end{theorem}

As discussed by \citet[Section~5]{DBLP:journals/pacmpl/Lee0A18}, this property can be further simplified when the exact result of the floating operation is not in the so-called ``subnormal'' range: the addictive error term $\delta'$ can be soundly removed, leading to a simplified $(1+\epsilon)$-property:
\begin{equation}
    x \circledast y = (x * y)(1+\delta),
    \quad
    \text{where }|\delta|\le\epsilon.
\label{equ:epsilon}    
\end{equation}

\subsection{Provable Quantization in \system}
\label{subsec:apply_pq}
\system applies provable quantization for an applicable operator $r(\cdot)$ with input $x$ in three steps: (\textit{i}) input quantization, (\textit{ii}) low-precision evaluation, and (\textit{iii}) exactness verification and, if needed, recalculation. 
In a nutshell, 
the input is first quantized into lower precision, and then the operator is evaluated in the lower precision, where $r(\cdot)$ is evaluated as $r_q(x)$. To verify the exactness of the quantization, we calculate the rounding error $\text{Err}(r_q(x))$ by the simplified $(1+\epsilon)$-property in eq. \eqref{equ:epsilon} and then check whether the result of $r(\cdot)$ may change with the rounding error. If it passes the verification, then we output $\text{Err}(r_q(x))$ as the final result; otherwise, we use the original precision of $x$ to recalculate $r(x)$. Note that we apply this technique to the entire input data $\mathbf{X} \in \mathbbm{R}^{n \times d}$, and only need to recalculate on the subset of $\mathbf{X}$ where the verification fails. \rv{Note that provable quantization does not apply to all operators, and we elaborate on the criteria in \S \ref{subsec:pq_criteria}.}

\subsection{Case Study: Neighbors Within Range}
\label{subsec:nwr}
We show the usage of provable quantization in \system on neighbors within range (\texttt{NWR}, one of the FOs of our programming model), a common step in many OD algorithms, e.g., LOF \cite{Breunig2000} and LOCI \cite{DBLP:conf/icde/PapadimitriouKGF03}. \texttt{NWR} identifies nearest neighbors within a preset distance threshold (usually a small number), which may be considered as a variant of $k$ nearest neighbors. More formally, given an input tensor $\bX:=[X_1, X_2, ...,X_n] \in \mathbb{R}^{n \times d}$ ($n$ samples and $d$ dimensions) and the  distance threshold $\phi$, \texttt{NWR} first calculates the pairwise distance among each sample via the \texttt{cdist} operator, yielding a distance matrix $\bD \in \mathbb{R}^{n\times n}$, where $\bD_{i,j}$ is the pairwise distance between $X_i$ and $X_j$. Then, each pairwise distance in $\bD$ is compared with $\phi$, and \texttt{NWR} outputs the indices of samples where $\bD_{i,j} \leq \phi$. As the pairwise distance calculation in \texttt{NWR} requires $O(n^2)$ space, provable quantization can provide significant GPU memory savings.

\texttt{NWR} meets the criterion we outline in \S \ref{subsec:apply_pq}, where eq.~\eqref{equ:epsilon} can be applied to estimate the rounding errors. Recall that the pairwise Euclidean distance between two samples is
\begin{equation}
\bD_{ij} = \norm{X_i-X_j}^2_2=\norm{X_i}_2^2+\norm{X_j}_2^2-2X_i^T X_j.
\label{eq:dist-decomp}
\end{equation}
We shall compute this in floating-point. Importantly, in our analysis to follow, the ordering of floating-point operations matters in determining rounding errors. To this end, we calculate distance using the right-most expression in eq.~\eqref{eq:dist-decomp} (note that we do \emph{not} first compute the difference $X_i-X_j$ and then compute its squared Euclidean norm). When calculating the first term in the RHS of eq.~\eqref{eq:dist-decomp} via floating-point operations, we get
\begin{align*}
    \text{fl}(\norm{X_i}_2^2)
    &\!=\!(X_{i,1}\otimes X_{i,1})\oplus(X_{i,2}\otimes X_{i,2})\oplus\cdots\oplus(X_{i,d}\otimes X_{i,d})\\
    &\!=\!(X_{i,1}^2(1\!+\!\delta_1))\oplus(X_{i,2}^2(1\!+\!\delta_2))\oplus\cdots\oplus(X_{i,d}^2(1\!+\!\delta_d)),
\end{align*}
where the second equality uses eq.~\eqref{equ:epsilon} and we note that the errors $\delta_1,\dots,\delta_d$ across the floating-point multiplications (for squaring) need not be the same (in fact, these need not be the same across samples $i=1,2,\dots,n$ but we omit this indexing to keep the equation from getting cluttered).

Next, by defining $x_{\max}\triangleq\max_{i\in\{1,\dots,n\},k\in\{1,\dots,d\}}|X_{i,k}|$ and recalling from Theorem~\ref{theorem:quantization} that each of $\delta_{1},\delta_{2},\dots,\delta_{d}$ above is at most $\epsilon$, we get
\begin{align*}
 \text{fl}(\norm{X_i}_2^2)
 & \!=\!(X_{i,1}^2(1\!+\!\delta_1))\oplus(X_{i,2}^2(1\!+\!\delta_2))\oplus\cdots\oplus(X_{i,d}^2(1\!+\!\delta_d))\\
 & \!\le\!\underbrace{[x_{\max}^{2}(1+\epsilon)]\oplus[x_{\max}^{2}(1+\epsilon)]\oplus\cdots\oplus[x_{\max}^{2}(1+\epsilon)]}_{d\text{ terms added via floating-point addition}}\\
 & \!\le\!d\cdot x_{\max}^{2}(1+\epsilon)^{1+\lceil\log_{2}d\rceil},
\end{align*}
where for the last step, $\log_2 d$ shows up since summation of $d$ elements in lower-level programming languages is implemented in a divide-and-conquer manner that reduces to $\lceil\log_2 d\rceil$ operations (there is still a ``$1+$'' term in the exponent for the floating-point multiplication/squaring). The rounding error is bounded as follows:
\begin{align*}
\text{Err}(\|X_{i}\|_{2}^{2}) & =\|X_{i}\|_{2}^{2}-\text{fl}(\|X_{i}\|_{2}^{2})\\
 & \le d\cdot x_{\max}^{2}-\text{fl}(\|X_{i}\|_{2}^{2})\\
 & \le\big|d\cdot x_{\max}^{2}-\text{fl}(\|X_{i}\|_{2}^{2})\big|\\
 & \le d\cdot x_{\max}^{2}[(1+\epsilon)^{1+\lceil\log_{2}d\rceil}-1].
\end{align*}
This same analysis can be used to bound the floating-point errors of the other terms in eq.~\eqref{eq:dist-decomp}. Overall, we get
\begin{equation}
    \text{Err}(\bD_{ij}) \leq 4d \cdot  x_{\text{max}}^2 [(1+\epsilon)^{1+\lceil\log_2 d\rceil+2} - 1],
\label{eq:nwr-check}
\end{equation}
where the ``$+2$'' shows up in the exponent due to the addition and subtraction in the RHS of~\eqref{eq:dist-decomp} that we compute in floating-point.

Inequality~\eqref{eq:nwr-check} provides a numerical way for checking whether a single entry $\bD_{ij}$ is within the range of $\phi$ as $|\bD_{ij}-\phi| > \text{Err}(\bD_{ij})$. More conveniently, we could scale the input  $\bX$ into the range of $[0, 1]$ before the distance calculation \cite{raschka2015python}, so that $x_{\text{max}} \leq 1$ and the implementation complexity can be further reduced. With this treatment, a large amount of GPU memory can be saved in \texttt{NWR} operations (see \S \ref{subsec:prov_quant_exp} for results).

\subsection{Applicability and Opportunities of Provable Quantization}
\label{subsec:pq_criteria}
Not all operators can benefit from provable quantization. To benefit from provable quantization, an operator needs to satisfy two criteria. First, the operator's output values cannot require a floating-point representation in the original precision, \rv{otherwise the exactness verification would require executing the operator in the original precision, resulting in no memory or time savings. For example, provable quantization is not applicable to \texttt{cdist} since its outputs are raw floating-point pairwise distances, and verifying its exactness requires calculating \texttt{cdist} in the original precision.}
Second, the performance gain in low-precision evaluation of the operator needs to be larger than the overhead of verification. Based on these two criteria, we mark the operators generally applicable for provable quantization by $\bigstar$ in Fig.~\ref{fig:operators}. Also see \S \ref{subsec:prov_quant_exp} for experimental results on this.
Although the design of provable quantization is motivated by unsupervised OD algorithms with extensive ranking and selecting operations, other ML algorithms can also benefit if they meet the above criteria.
\rv{For OD algorithms in which provable quantization does not apply, they could still benefit from \system's other optimizations such as automatic batching (\S \ref{sec:batching}).}

\section{Automatic Batching and Multi-GPU Support}
\label{sec:batching}

\begin{figure}[!t] 
\centering
  \includegraphics[width=0.4\textwidth]{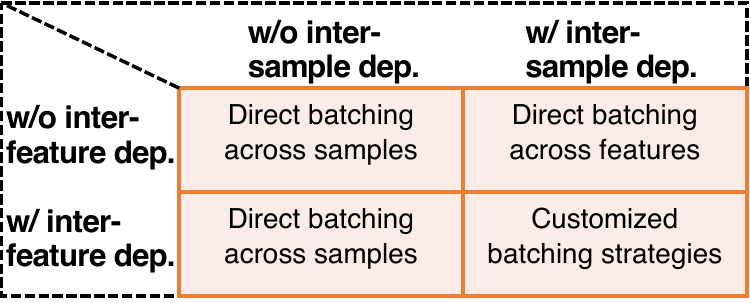}
  \vspace{-0.1in}
  \caption{\system applies automatic batching to operators without inter-sample or inter-feature dependency, and uses customized batching strategies for operators with both data dependencies.
  }
  \vspace{-0.1in}
  \label{fig:batching_mechanism}
\end{figure}

\MyPara{Motivation.} Unlike CPU nodes with up to terabytes of DRAM, today's GPU nodes face much more stringent memory limitations~\cite{min2020emogi,gera2020traversing,min2021large,lee2021art}---modern GPUs are mostly equipped with 4-40 GB of on-device memory \cite{meng2017training,DBLP:conf/heart/SvedinCCJP21}. Out-of-memory (OOM) errors have thus become common in GPU-based systems for machine learning tasks \cite{DBLP:conf/sigsoft/GaoLZLZLY20}. 
To overcome this challenge, we design an {\em automatic batching} mechanism to decompose memory-intensive BTOs into multiple batches, which are executed on GPUs in a pipeline fashion. Automatic batching also allows \system to equally distribute OD computation across multiple GPUs.
\rv{\system applies different mechanisms to decompose an operator into batches based on its data dependency.
Specifically, an operator has {\em inter-sample dependency} if the computation of each sample requires accessing other samples, such as \texttt{cdist}.
On the other hand, an operator has {\em inter-feature dependency} if the computation of each feature depends on other features, such as \texttt{Feat. sampler}.
Fig. \ref{fig:batching_mechanism} summarizes \system batching mechanisms for operators with and without these dependencies.}

\begin{figure}[!t] 
\centering
  \includegraphics[width=0.45\textwidth]{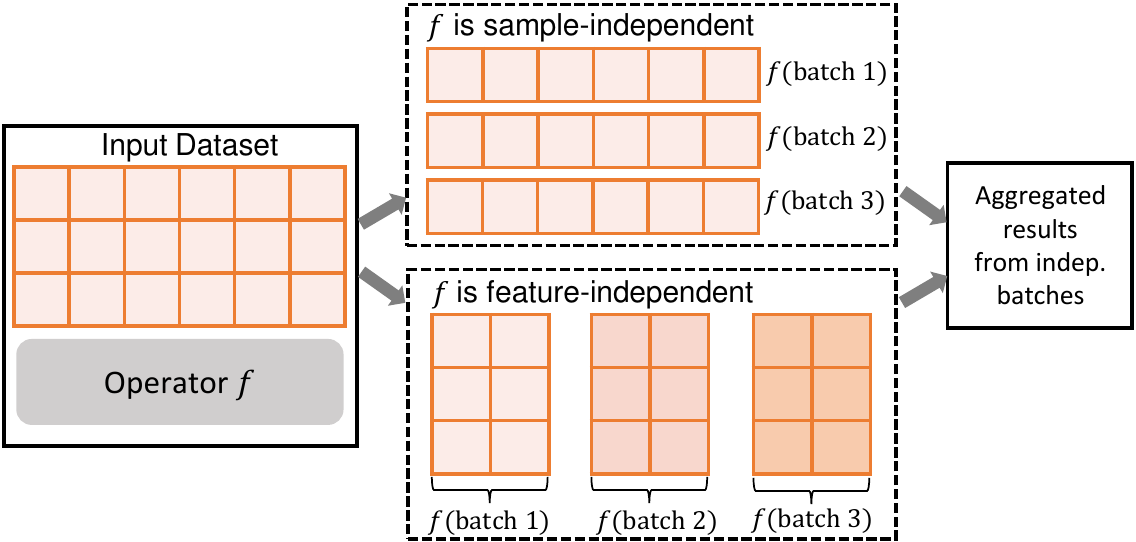}
  \vspace{-0.1in}
  \caption{Direct batching with independence assumption creates batches along the sample or feature index.}
 \label{fig:trivial_batching}
 \vspace{-0.1in}
\end{figure}

\noindent \rv{\MyPara{Direct batching.}} \system automatically decomposes an operator into small batches if the operator is: (\textit{i}) \textit{sample-independent}: the estimation of each sample is independent, or (\textit{ii}) \textit{feature-independent}: the contribution of each feature is independent. When either condition holds, \system directly partitions the operator into multiple batches by splitting along the sample or feature dimension, computes these batches in a pipeline fashion, and aggregates individual batches to produce the final output, as shown in Fig.~\ref{fig:trivial_batching}.
For instance, \texttt{topk} is a sample-independent operator. For an input tensor $\bX \in \mathbb{R}^{n\times d}$, {\tt topk} outputs the indices of the largest $k$ values for each sample (i.e., row), resulting in an output tensor $\bI_{\bX}:= \texttt{topk}(\bX, k) \in \mathbb{R}^{n\times k}$. Therefore, we could split $\bX$ into batches with full features, each with $b$ samples (i.e.,, $\{\bX_{1}, \bX_{2}, ...\}  \in \mathbb{R}^{b\times d}$). As another example, \texttt{Histogram} outputs the frequencies of each feature's values, which is feature-independent. Thus, we could partition $\bX$ into blocks of $b$ features, e.g., $\{\bX_{1}, \bX_{2}, ...\} \in \mathbb{R}^{n\times b}$.

\noindent \rv{\MyPara{Customized batching.}
For operators with {\em both} inter-sample and inter-feature dependencies, the computation of each data point involves all samples and features, and therefore cannot be directly decomposed into batches along the same or feature dimension. \system provides customized batching strategies for these operators. For example, to automatically batch \texttt{cdist}, \system uses the approach introduced in \citet{neeb2016distributed}, which splits an input dataset across samples and calculates the pairwise distance of {\em each pair of split} in batches.
}

\begin{figure}[!t] 
\centering
  \includegraphics[width=0.45\textwidth]{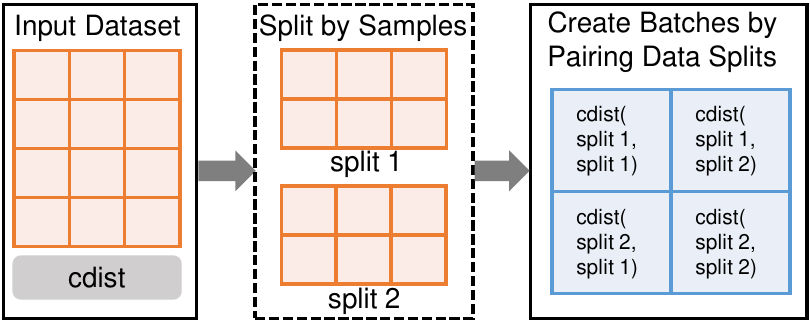}
  \vspace{-0.1in}
  \caption{Customized batching solution for cdist in \system.}
    \vspace{-0.2in}
 \label{fig:cdist_simple}
\end{figure}

\subsection{Sequential Batching and Operator Fusion}
\label{subsec:operator_fusion}
\MyPara{Simple concatenation.} Since BTOs are independent from each other, executing a sequence of BTOs in batches is straightforward, i.e., simply feed the output of a batch operator as an input to another one. For example, \texttt{$k$NN} (see Fig.~\ref{fig:operators}) finds the $k$ nearest neighbors by first calculating pairwise distances of input samples via the \texttt{cdist} BTO and then returns the index of $k$ items with the smallest distance via the \texttt{topk} BTO.
Thus, \texttt{$k$NN} batching is achieved by running \texttt{cdist} and \texttt{topk} sequentially, where each uses automatic batching and the output of the former is the input of the latter. 
\rv{Note that simple concatenation applies to all BTOs and FOs as the default choice.}

\begin{figure} 
	\subfloat[Simple concatenation: \texttt{topk} is invoked on the full result of \texttt{cdist}---we need to communicate the large distance matrix $\bD$.]{%
		\includegraphics[clip,width=1.02\columnwidth]{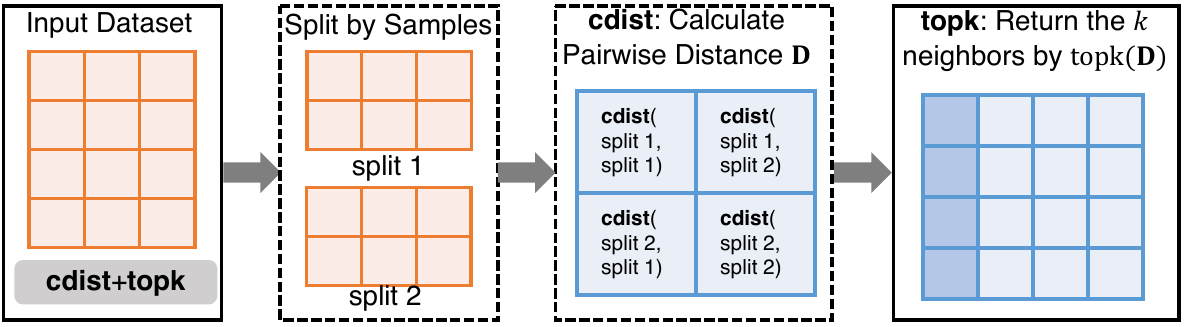}%
	}
	
	\subfloat[Operator fusion: \texttt{topk} is directly invoked on the batch result of \texttt{cdsit}, preventing the communication of distance matrix $\bD$.] {%
		\includegraphics[clip,width=1.02\columnwidth]{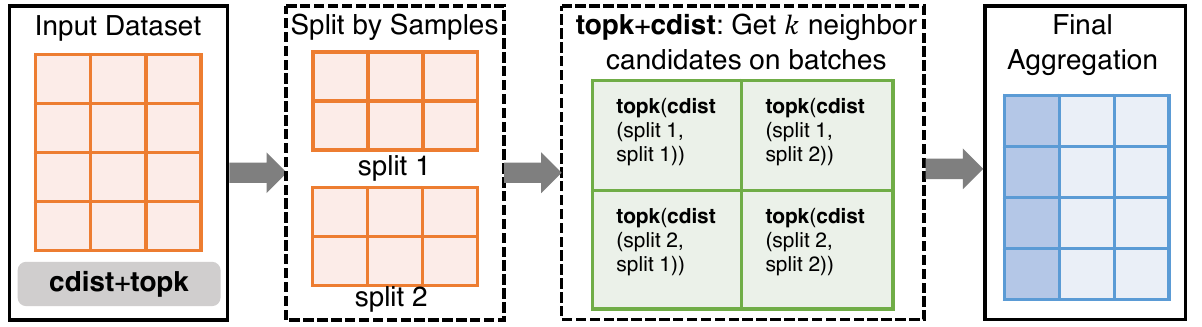}%
	}
	\caption{The comparison of automatic batching for $k$NN between simple concatenation and operator fusion. The latter has better scalability by not creating and moving large distance matrix $\bD$.}
	\label{fig:knn_direct_fusion}
 \vspace{-0.2in}
\end{figure}

\MyPara{Operator fusion.} Although the simple concatenation discussed above is straightforward, a closer look unlocks deeper optimization opportunities in automatic batching with a sequence of operators. Notably, the output of \texttt{$k$NN} is the indices of the $k$ nearest neighbors of an input dataset, where the pairwise distance generated by \texttt{cdist} is only used in an intermediate step but not returned.
If we could prevent moving this large distance matrix between operators (i.e., \texttt{cdist} and \texttt{topk}), space efficiency can be improved.
In deep learning systems, {\em operator fusion} is a common optimization technique to fuse multiple operators into a single one in a computational graph \cite{DBLP:conf/nips/NeubigGD17,DBLP:conf/iclr/LooksHHN17,DBLP:conf/osdi/WangZGMTZLRCJ21,boehm2018optimizing,menon2017relaxed}. 
Fig.~\ref{fig:knn_direct_fusion} compares simple concatenation (subfigure a) and operator fusion (subfigure b) on \texttt{$k$NN}. Specifically, the latter executes the \texttt{topk} BTO on the \texttt{cdist} BTO's individual batches separately rather than running \texttt{topk} on the full distance matrix outputted by \texttt{cdist}. Note that the global $k$ nearest neighbors (of the full dataset) can be identified from the $k$ local neighbor candidates from batches in the final aggregation, so the result is still exact. This prevents moving the entire $n \times n$ distance matrix between operators, which often causes OOM.
\rv{
\system uses a {\em rule-based} approach to opportunistically fusing operators to reduce the kernel launch overhead and data transfers between CPUs and GPUs.
\system provides an interface that allows users to add fusion rules for new OD operators.
Appx. \ref{appx:fusion_study} provides a case study on the effectiveness of operator fusion.}

\subsection{Multi-GPU Support}
\label{subsec:multigpu}
\rv{To further reduce the execution time of OD algorithms on GPUs, \system also supports multi-GPU execution, which is important for time-critical OD applications and has been widely used for other data-intensive applications such as graph neural networks~\cite{DBLP:conf/mlsys/JiaLGZA20}. }
Intuitively, if there is only one GPU, \system iterates across multiple batches sequentially and aggregates the results.
When multiple GPUs are available, we could achieve better performance by executing OD computations concurrently on multiple \rv{homogeneous} GPUs.
Specifically, \system first applies automatic batching to an underlying task---multiple subtasks are created and assigned to available GPUs.
\system creates a {\em subprocess} for each available GPU to execute the assigned subtasks and a {\em shared global container} to store the results returned from each GPU. 
Since we equally distribute subtasks across GPUs, we deem the runtime of each GPU is close. 
Once all the subtasks are complete, the final output is generated by aggregating the results in the global container. For example, automatically batching \texttt{cdist} in Fig.~\ref{fig:cdist_simple} leads to 4 subtasks, each of which calculates the pairwise distances for a pair of splits (denoted as blue blocks in the figure).
Each of the four available GPUs executes an assigned subtask and sends the \texttt{cdist} results to the global container. Finally, the full \texttt{cdist} result is obtained by aggregating the intermediate results in the global container. Note that the multi-GPU execution is at the operator level (e.g., \texttt{cdist}). \rv{\S \ref{subsec:multi_gpu} evaluates \system's scalability across multiple GPUs.}.

\section{Experimental Evaluation}
\label{sec:experiments}
Our experiments answer the following questions:
\begin{enumerate}[leftmargin=*]
\setlength\itemsep{0.025in}
\item  Is \system more efficient (in time and space) than SOTA \textit{CPU-based} OD system (i.e., PyOD) and selected \textit{GPU} baselines? (\S \ref{subsec:end_2_end})
\item How scalable is \system while handling more and more data? (\S \ref{subsec:scalability})
\item  How effective are provable quantization and automatic batching), in comparison to PyTorch implementation? (\S \ref{subsec:prov_quant_exp} \& \ref{subsec:batching_exp})
\item  How much performance gain can \system achieve on the multiple GPUs? (\S \ref{subsec:multi_gpu})
\end{enumerate}

\subsection{Implementation and Environment}
\label{subsec:implement}
\system is implemented on top of PyTorch \cite{paszke2019pytorch}. We extend PyTorch in the following aspects to support efficient OD. First, we implement a set of BTOs and FOs (see Fig.~\ref{fig:operators}) for fast tensor operations in OD. Second, for operators that support provable quantization  (see \S\ref{sec:quantization}) and batching (see \S\ref{sec:batching}), we create corresponding versions of them to improve scalability. Additionally, we enable specialized multi-GPU support in \system by leveraging PyTorch's \texttt{multiprocessing}. 
The usage and APIs of the open-sourced system can be found in \S \ref{sec:api}.

\noindent \rv{\MyPara{Adding new operators}. 
In addition to the BTOs and FOs listed in Fig. \ref{fig:operators}, users can add new operators in \system by defining the operator's interface (i.e., the input and output tensors of the operator) and providing an implementation of the operator in PyTorch.
This implementation will be used by \system to decompose the operator into PyTorch's tensor algebra primitives and execute these primitives in parallel on multiple GPUs.
For operators that do not have inter-sample {\em or} inter-feature dependency (see \S \ref{sec:batching}), \system automatically decomposes the operator's computation into multiple batches. For operators that involve both inter-sample {\em and} inter-feature dependencies, \system require users to provide a customized strategy to decompose the operator into batches.
}

\noindent \rv{\MyPara{Implementing new OD algorithms}. One notable characteristic of OD is that most algorithms involve only straightforward computation, which can be decomposed into 2-3 BTOs and FOs. Therefore, we expect that implementing new OD algorithms in \system should only involve low cognitive complexity. Appx. \ref{appx:ecod} demonstrates an implementation of a recent ECOD~\cite{li2022ecod} detection algorithm in \system in less than ten lines of code.
} 

\noindent \MyPara{Experimental setup.} All the experiments were performed on an Amazon EC2 cluster with an Intel Xeon E5-2686 v4 CPU, 61GB DRAM, and an NVIDIA Tesla V100 GPU. For the multi-GPU support evaluation, we extend it to multiple NVIDIA Tesla V100 GPUs with the same CPU node.

\subsection{Datasets, Baselines, and Evaluation Metrics}
\label{subsec:data_base_metrics}
\MyPara{Datasets.} \rv{Table \ref{table:data} shows the 11 real-world benchmark datasets used in this study, which are widely evaluated in OD research \cite{campos2016evaluation,zhao2019lscp,ruff2019deep,DBLP:conf/icdm/LiZBIH20} and available in the latest ADBench\footnote{Datasets available at ADBench: \url{https://github.com/Minqi824/ADBench}} \cite{han2022adbench}.}
Given the limited size of real-world OD datasets, we also build data generation function in \system to create larger synthetic datasets (up to 1.5 million samples) to evaluate the scalability of \system (see \S \ref{subsec:scalability} for details).

\noindent \MyPara{OD algorithms and operators.} Throughout the experiments, we compare the performance of five representative but diverse OD algorithms across different systems (see \S \ref{subsec:od_algorithms}): 
proximity-based algorithms including LOF \cite{Breunig2000}, ABOD \cite{kriegel2008angle}, and $k$NN \cite{Angiulli2002fast}; statistical method HBOS \cite{goldstein2012histogram}, and linear model PCA \cite{shyu2003novel}. We also provide an operator-level analysis on selected BTOs and FOs to demonstrate the effectiveness of certain techniques.

\noindent \MyPara{Evaluation metrics.} Since \system and the baselines do not involve any approximation, the output results are exact and consistent across systems. Therefore, we omit the accuracy evaluation, and compare the wall-clock time and GPU memory consumption as measures of time and space efficiency.

\noindent \MyPara{Baselines.} As discussed in Section \ref{sec:related_work}, there is no existing GPU system that covers a diverse group of OD algorithms (not even the above five algorithms) for a fair comparison. Therefore, we use the SOTA comprehensive system PyOD \cite{DBLP:journals/jmlr/ZhaoNL19} as a \textit{CPU} baseline in \S \ref{subsec:end_2_end} and \ref{subsec:scalability}, which is deeply optimized with JIT compilation and parallelization. Regarding \textit{GPU} baselines, we compare two representative OD algorithms (i.e., $k$NN-CUDA \cite{DBLP:journals/tpds/AngiulliBLS16} and LOF-CUDA \cite{alshawabkeh2010accelerating}) that have GPU support in \S \ref{subsec:end_2_end}, and direct implementation of operators in PyTorch in \S \ref{subsec:prov_quant_exp}, \ref{subsec:batching_exp}, and \ref{subsec:multi_gpu}. Note that the implementation of $k$NN-CUDA and LOF-CUDA are not open-sourced, so we follow the original papers to implement.

\begin{table} 
	\centering
	\caption{Real-world OD datasets used in the experiments. To demonstrate the results on larger datasets, we also create and use synthetic datasets throughout the experiments. 
 } 
	\label{table:data} 
	\begin{tabular}{l | llll } 
		\toprule 
		\textbf{Dataset} & \textbf{Pts (\textbf{\textit{n}})} & \textbf{Dim (\textbf{\textit{d}})} & \textbf{\% Outlier}\\
		\midrule
		musk          & 3,062  & 166  & 3.17  \\
		speech        & 3,686  & 400  & 1.65 \\
		mnist         & 7,603  & 100  & 9.21  \\
		mammography   & 11,183 & 6    & 2.32 \\
		ALOI          & 49,534 & 27   & 3.04 \\
		fashion-mnist & 60,000 & 784  & 10    \\
		cifar-10      & 60,000 & 3072 & 10    \\
            celeba        & 202,599&	39	 & 2.24\\
            fraud         & 284,807&	29&	0.17\\ 
            census        & 299,285&	500& 6.20\\
            donors        & 619,326&	10&	5.93\\

		\bottomrule
	\end{tabular}
	
\end{table}

\begin{figure}[!tp] 
\centering

\subfloat[ALOI (49,534 samples with 27 features)]{%
  \includegraphics[clip,width=0.85\columnwidth]{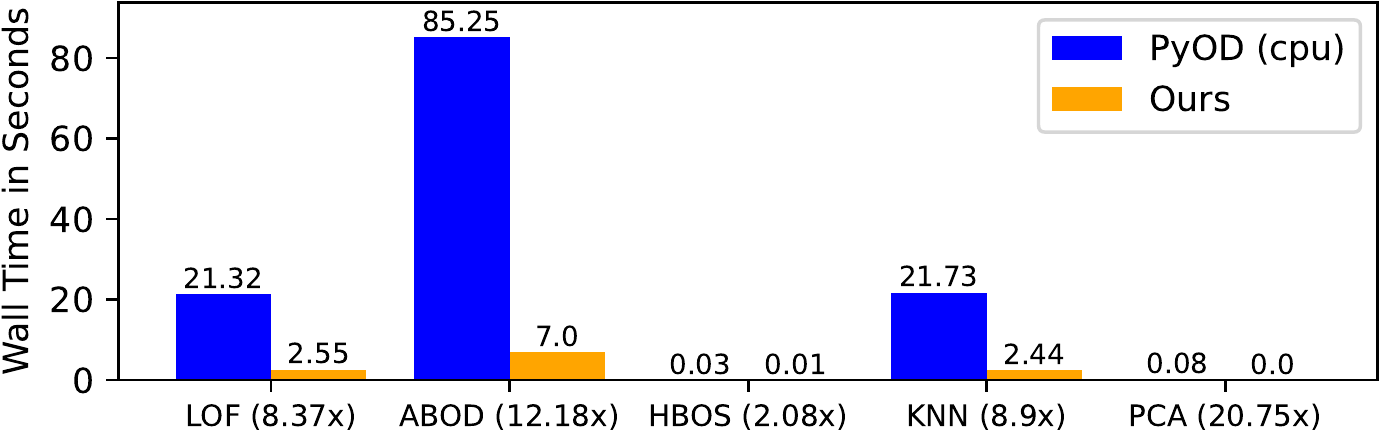}%
}
\vspace{0.01in}

\subfloat[cifar-10 (60,000 samples with 784 features)]{%
  \includegraphics[clip,width=0.85\columnwidth]{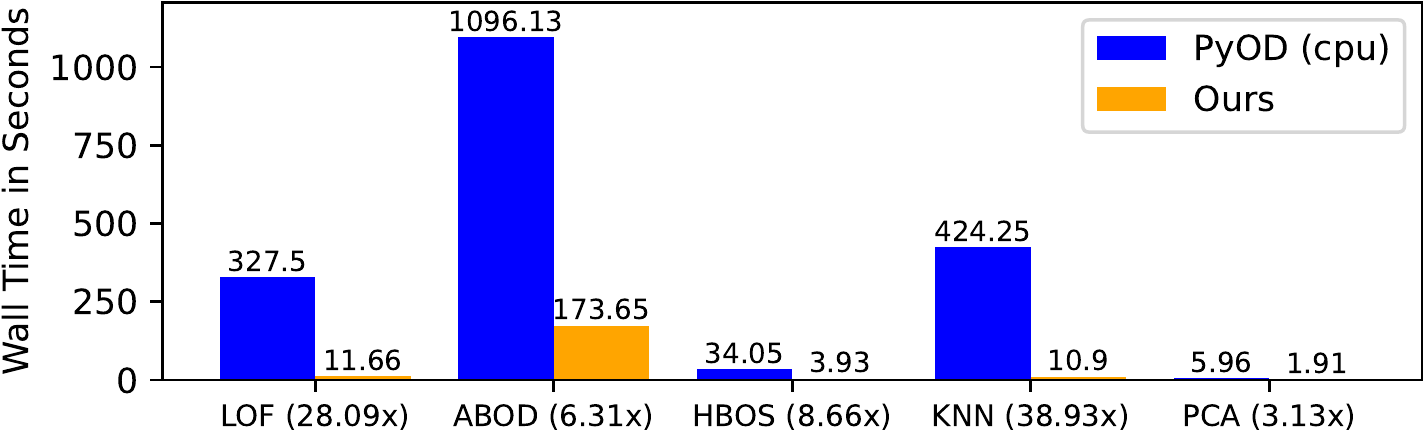}%
}
\vspace{0.01in}

\subfloat[fashion-mnist (60,000 samples with 3,072 features)]{%
  \includegraphics[clip,width=0.85\columnwidth]{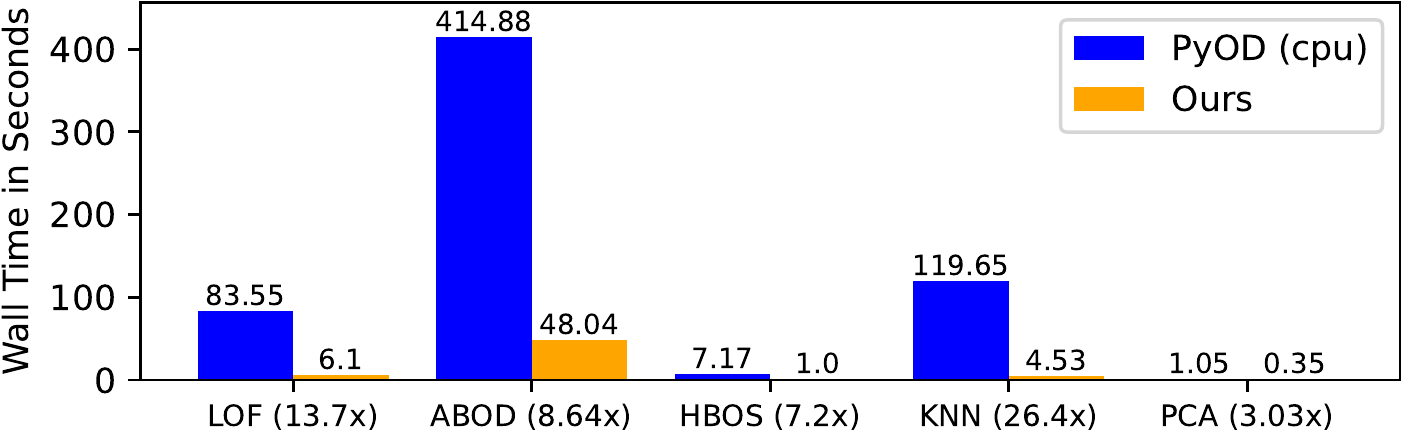}%
}
\vspace{0.01in}

\subfloat[celeba (202,599 samples with 39 features) ]{%
  \includegraphics[clip,width=0.85\columnwidth]{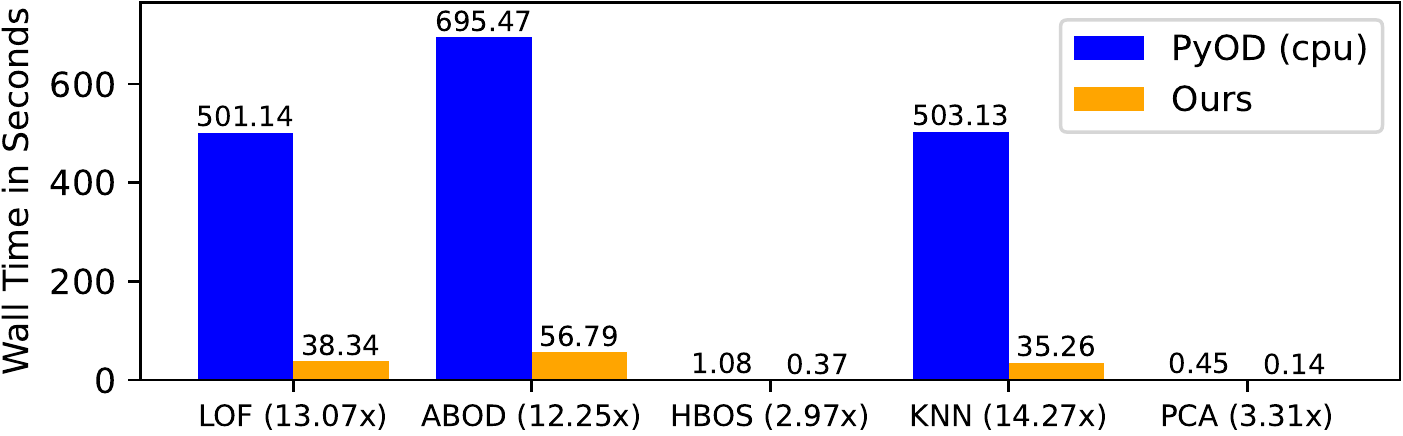}%
}  
\vspace{0.01in}

\subfloat[fraud (284,807 samples with 29 features) ]{%
  \includegraphics[clip,width=0.85\columnwidth]{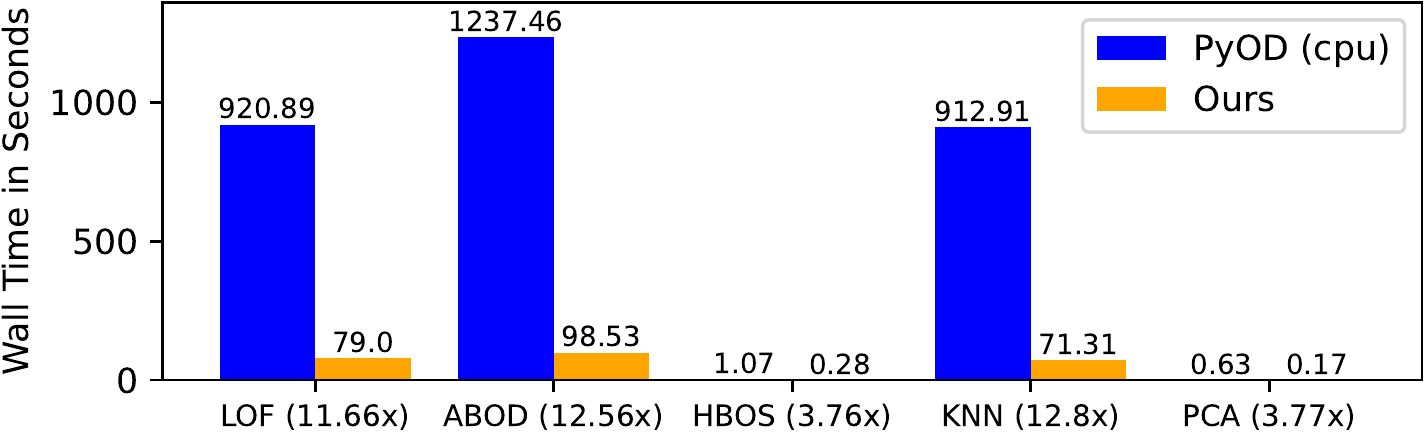}%
} 
\vspace{0.01in}

\subfloat[census (299,285 samples with 500 features)]{%
  \includegraphics[clip,width=0.85\columnwidth]{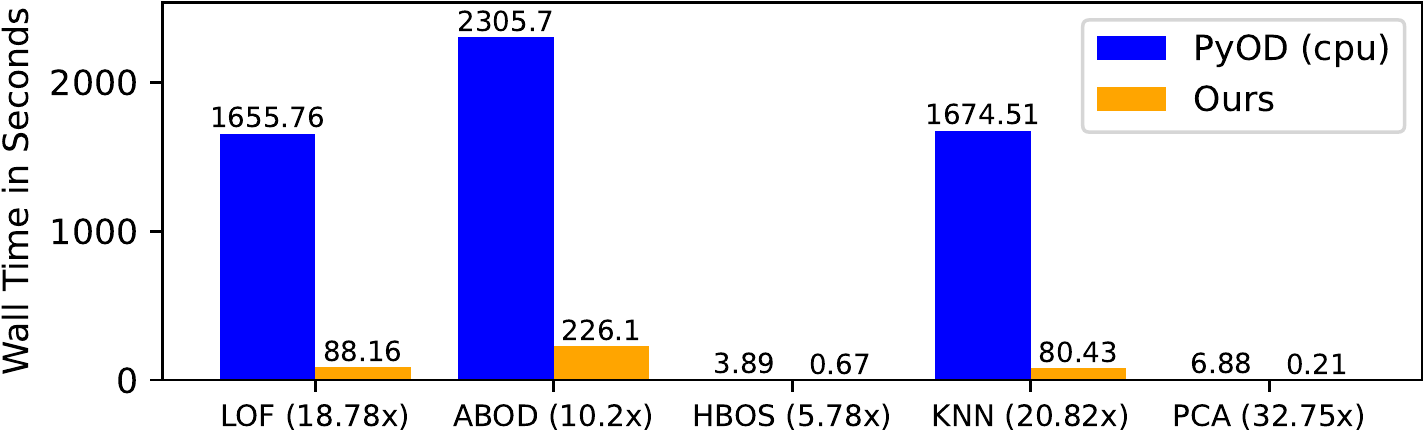}%
}
\vspace{0.01in}

\subfloat[donors (619,326 samples with 10 features) ]{%
  \includegraphics[clip,width=0.85\columnwidth]{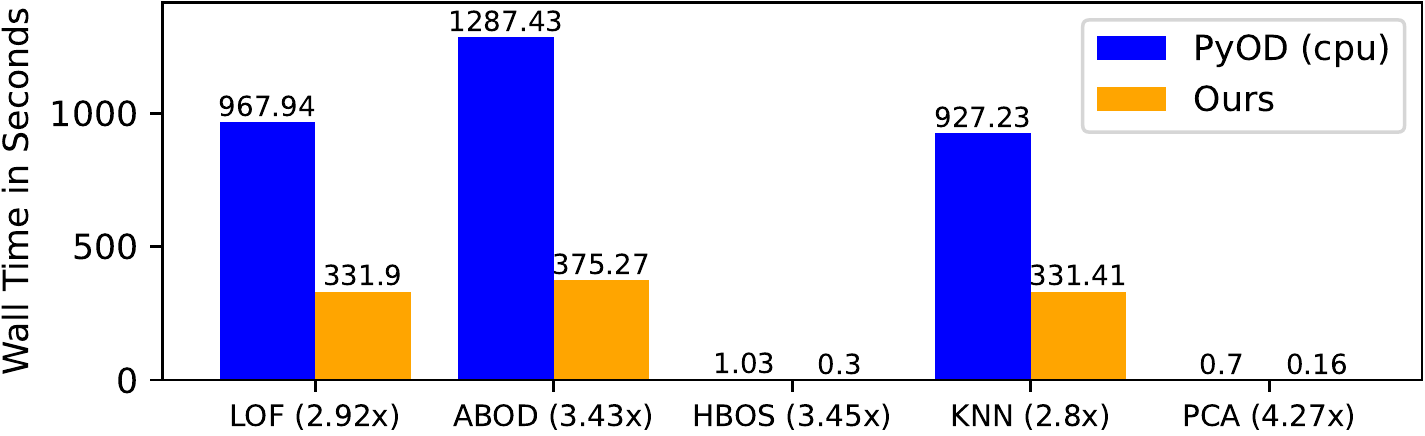}%
} 
\vspace{-0.15in}
\caption{\rv{Runtime comparison between PyOD and \system in seconds on both real-world and synthetic datasets (see Appx. Fig \ref{fig:full_system_synthetic} for synthetic data results). \system significantly outperforms PyOD in all w/ much smaller runtime, where the speedup factor is shown in parenthesis by each algorithm. On avg., \system is 10.9$\times$ faster than PyOD (up to 38.9$\times$)}.}
\label{fig:full_system_evaluation}
\end{figure}

\subsection{End-to-end Evaluation}
\label{subsec:end_2_end}
\textbf{\system is significantly faster than the leading CPU-based system}. \rv{We first present the runtime comparison between \system and PyOD in Fig.~\ref{fig:full_system_evaluation} using seven real datasets (ALOI, fashion-mnist, and cifar-10)} and Appx. Fig. \ref{fig:full_system_synthetic} with three synthetic datasets (where Synthetic 1 contains 100,000 samples, Synthetic 2 contains 200,000 samples, and Synthetic 3 contains 400,000 samples (all are with 200 features).
\rv{The results show that \system is on average 10.9$\times$ faster than PyOD on the five benchmark algorithms (13.0$\times$, 15.9$\times$, 9.3$\times$, 7.2$\times$, and 8.9$\times$ speed-up on LOF, $k$NN, ABOD, HBOS, and PCA, respectively).}
For proximity-based algorithms, a larger speed-up is observed for datasets with a higher number of dimensions: LOF and $k$NN are 28.1$\times$ and 38.9$\times$ faster on cifar-10 with 3,072 features. This is expected as GPUs are well-suited for dense tensor multiplication, which is essential in proximity-based methods. Separately, a larger improvement can be achieved for HBOS and PCA on datasets with larger sample sizes. For instance, HBOS is 11.83$\times$ faster on Synthetic 1 (100,000 samples), while the speedup is 17.16$\times$ on Synthetic 2 (200,000 samples). This is expected as HBOS treats each feature independently for density estimation on GPUs, so a large number of samples with a small number of features should yield a 
significant speed-up. In summary, all the OD algorithms tested are significantly faster in \system than in the SOTA PyOD system, with the precise amount of speed improvement varying across algorithms. \rv{Appx. \ref{appx:time_analysis} shows that GPU computation takes most of the run time for various OD algorithms, explaining the significant time reduction by \system. We also provide ablation studies to evaluate provable quantization and automatic batching in Appx. \ref{appx:ablations}.}

\noindent \textbf{\system can handle larger datasets than the GPU baselines}. Due to the absence of GPU systems that support all five OD algorithms, we specifically compare the performance of \system to specialized GPU algorithms $k$NN-CUDA \cite{DBLP:journals/tpds/AngiulliBLS16} and LOF-CUDA \cite{alshawabkeh2010accelerating}.  Table \ref{table:gpu} shows that \system with 8-GPUs outperforms in all three datasets due to the multi-GPU support, which enables it to handle data more efficient than the baselines and \system with a single GPU. By focusing on the use of a single GPU, we find that \system is faster or on par with both baselines due to provable quantization (e.g., 33.13\% speedup to LOF-CUDA). 
Also note that the GPU baselines face the out-of-memory issue (OOM) on large datasets (e.g., the last row of the table with 2,000,000 samples), while \system can still handle it due to automatic batching. In comparison to these specialized GPU baselines, \system does not only provide more coverage of diverse algorithms, but also yields better efficiency and scalability.

\begin{table}[!tp] 
\centering
	\caption{Runtime comparison among selected GPU baselines ($k$NN-CUDA \cite{DBLP:journals/tpds/AngiulliBLS16} and LOF-CUDA \cite{alshawabkeh2010accelerating}; neither supports multi-GPU directly), and \system (single GPU) and \system-8 (8 GPUs). The first column shows three synthetic datasets with an increasing number of samples (100 dimensions), and the most efficient result is highlighted in bold for each setting. \system-8 outperforms in all cases due to multi-GPU support, while \system with a single GPU is faster or on par with the baselines. Note that the GPU baselines run out-of-memory (OOM) on large datasets (e.g., the last row), while \system does not.} 

	\footnotesize
	\scalebox{0.96}{
    \begin{tabular}{l|lll|lll}
    \toprule
    \textbf{Dataset} & \textbf{$k$NN-CUDA} & \textbf{\system} & \system-8 & \textbf{LOF-CUDA} & \textbf{\system}&  \system-8\\
    \midrule
    500,000              & 205.33       & 208.84          &\textbf{28.59}  & 312.55 & 209& \textbf{28.64}          \\
    1,000,000              & 850.12      & 827         & \textbf{112.35}  & OOM  & 819 &\textbf{113.72}\\
    2,000,000              & OOM      & 3173.39           & \textbf{430.29}           & OOM & 3174.05 & \textbf{434.18}\\
    \bottomrule
    \end{tabular}
    }
	\label{table:gpu} 
 \vspace{-0.2in}

\end{table}

\begin{figure*}[!htp] 
\centering
\subfloat[LOF (quadratic)]{%
  \includegraphics[clip,width=0.45\columnwidth]{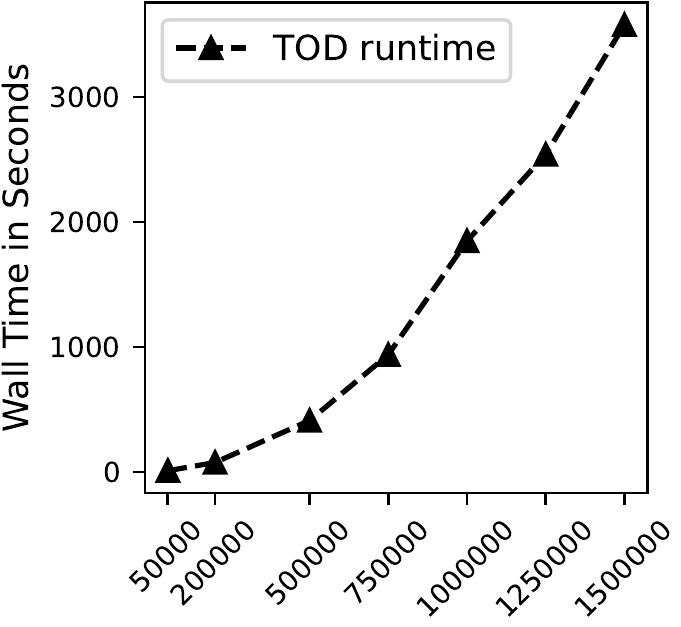}%
}
\subfloat[ABOD (quadratic)]{%
  \includegraphics[clip,width=0.41\columnwidth]{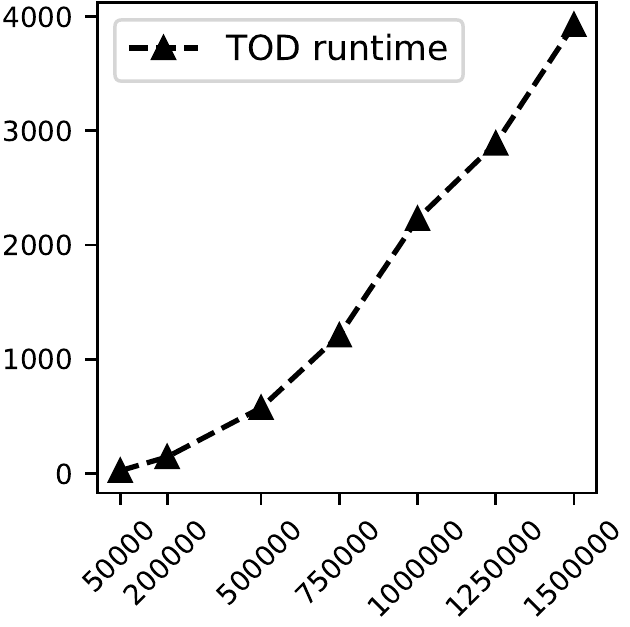}%
}
\subfloat[HBOS (linear)]{%
  \includegraphics[clip,width=0.41\columnwidth]{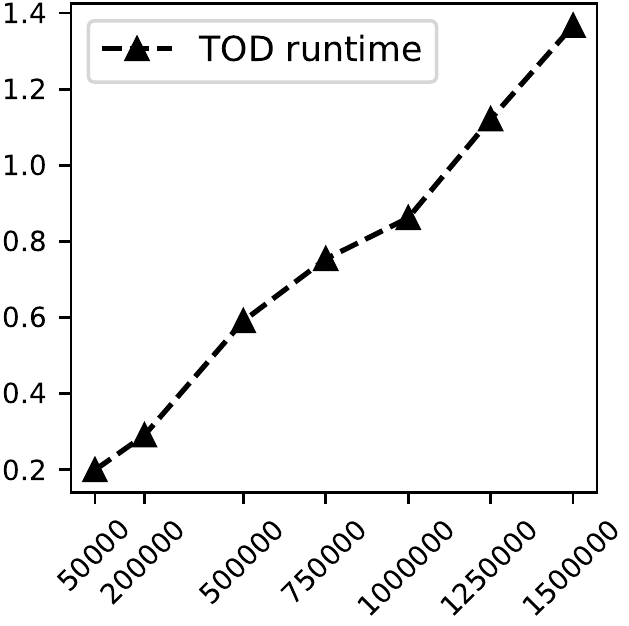}%
}  
\subfloat[$k$NN (quadratic)]{%
  \includegraphics[clip,width=0.41\columnwidth]{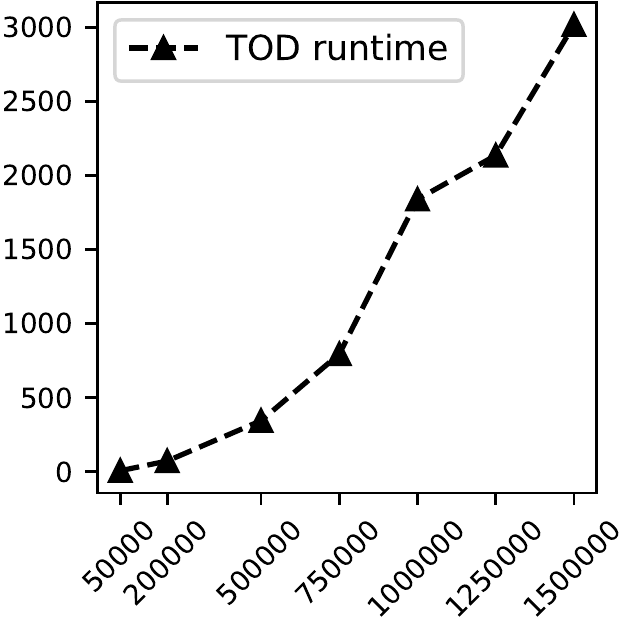}%
}  
\subfloat[PCA (linear)]{%
  \includegraphics[clip,width=0.41\columnwidth]{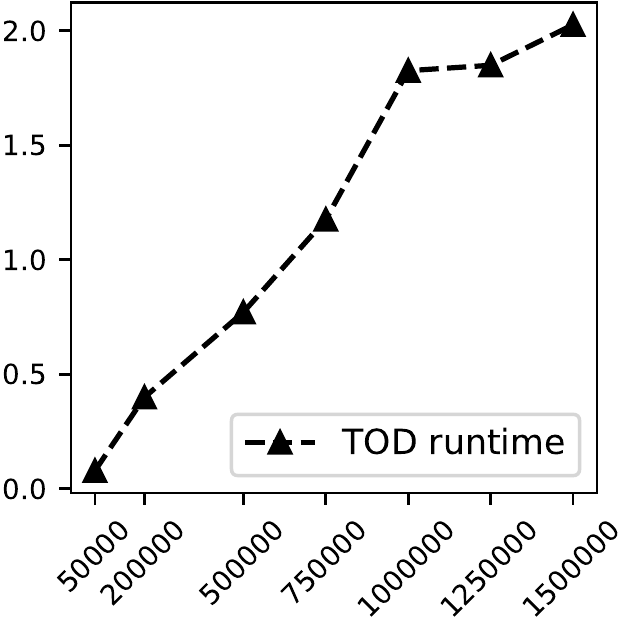}%
}  
\vspace{-0.1in}
\caption{\rv{Scalability plot of selected algorithms in \system, where it scales well with an increasing number of samples.}}
\label{fig:scalability}
\end{figure*}

\subsection{Scalability of \system}
\label{subsec:scalability}
We now gauge the scalability of \system on datasets of varying sizes, including ones larger than fashion-mnist and cifar-10.
In Fig.~\ref{fig:scalability}, we plot \system's runtime with five OD algorithms on the synthetic datasets with sample sizes ranging from 50,000 to 1,500,000 (all with 200 features). To the best of our knowledge, none of the existing comprehensive OD systems can handle datasets with more than a million samples within a reasonable amount of time \cite{aggarwal2015outlier,zhao2021suod}, as most of the OD algorithms are associated with quadratic time complexity.
\rv{Fig.~\ref{fig:scalability} shows that \system can process million-sample OD datasets within an hour, providing a scalable approach to deploying OD algorithms in many real-world tasks. 
}

\begin{figure*}[t] 
\centering
\subfloat[\texttt{Neighbor within range} (real-world datasets)]{%
  \includegraphics[clip,width=1\columnwidth]{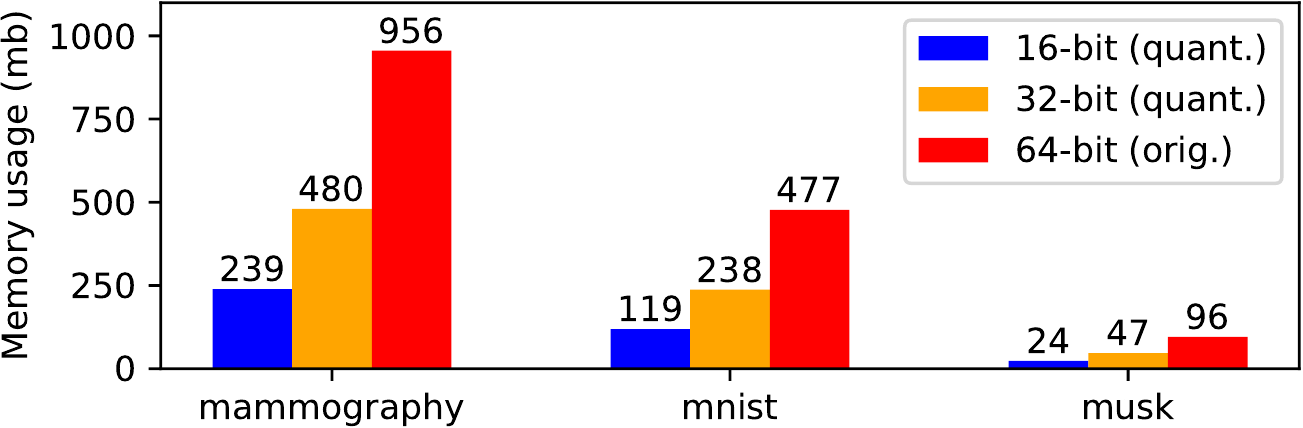}%
}
\hspace{0.15in}
\subfloat[\texttt{Neighbor within range} (synthetic datasets)]{%
  \includegraphics[clip,width=1\columnwidth]{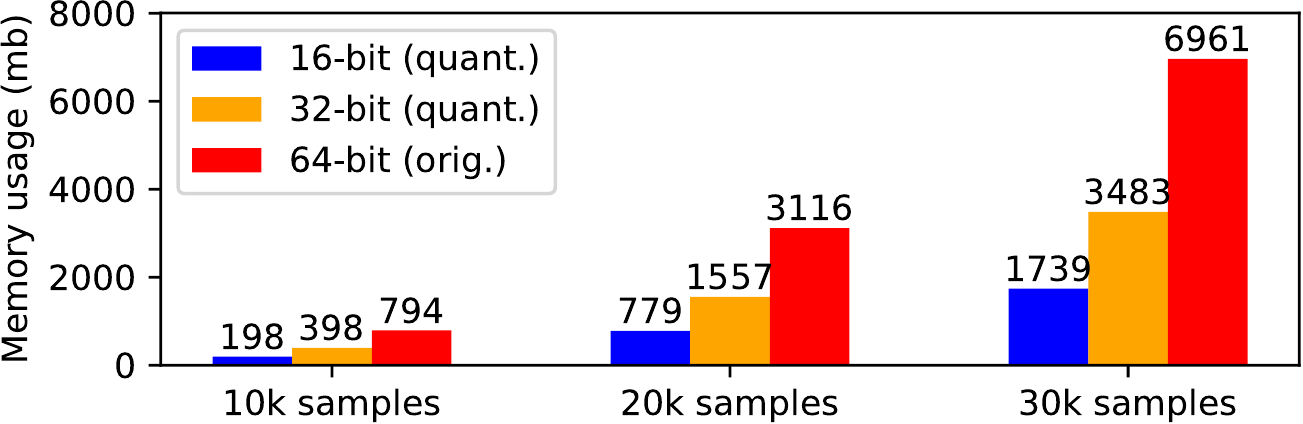}%
}

\subfloat[\texttt{topk} (real-world datasets)]{%
  \includegraphics[clip,width=1\columnwidth]{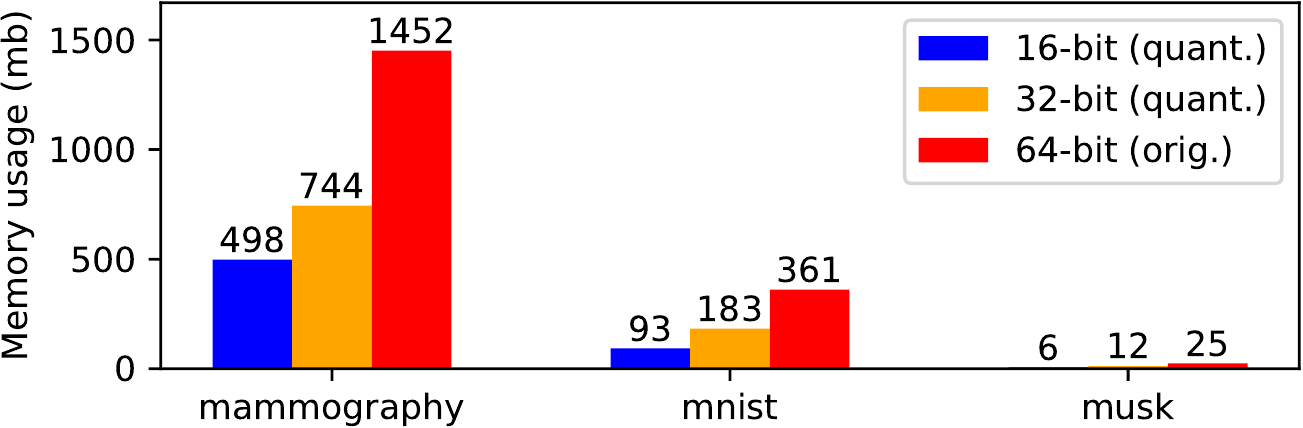}%
} 
\hspace{0.15in}
\subfloat[\texttt{topk} (synthetic datasets)]{%
  \includegraphics[clip,width=1\columnwidth]{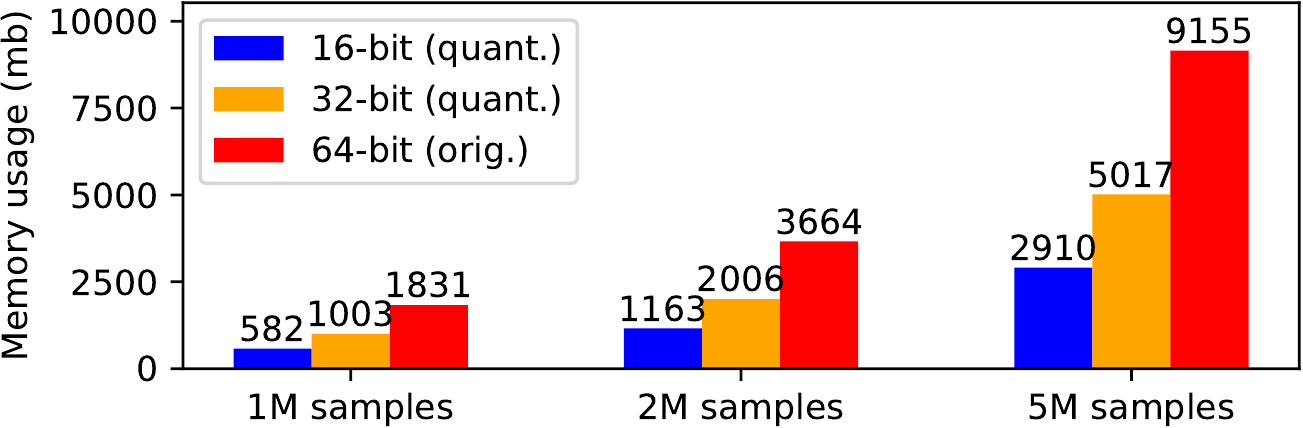}%
}
\vspace{-0.1in}
\caption{GPU memory consumption comparison between using provable quantization (16-bit and 32-bit) and the full precision (64-bit). Clearly, provable quantization leads to significant memory consumption saving on \texttt{nwr} and \texttt{topk}.
}
\label{fig:quantization_evaluation}
\end{figure*}

\begin{table}[!t]
\caption{Comparison of operator runtime (in seconds) with provable quantization (i.e., 16-bit and 32-bit) and without quantization (i.e., 64-bit) for \texttt{nwr} and \texttt{topk}. The best model is highlighted in bold (per column), where provable quantization in 16-bit outperforms the rest in most cases.}
\centering
\label{table:quantization}
    \begin{subtable}[!htb]{0.45\textwidth}
        \centering
        \begin{tabular}{l|llllll}
        \toprule
\textbf{Prec.}       & mammog. & mnist & musk & 10k  & 20k  & 30k \\
\midrule
\textbf{16-bit} & \textbf{2.66}        & \textbf{0.54}  & \textbf{0.06} & \textbf{0.87} & \textbf{3.02} & \textbf{6.56}   \\
\textbf{32-bit} & 2.92        & 1.51  & 0.09 & 2.57  & 10.09  & 22.51\\
\textbf{64-bit} & 3.31        & 1.53  & 0.09 & 3.49 & 12.7  & 27.32 \\
\bottomrule
\end{tabular}
       \caption{For \texttt{nwr}, 16-bit provable quantization outperforms in all}
       \label{table:quant_nwr}
    \end{subtable}
    \hfill
    \begin{subtable}[!ht]{0.45\textwidth}
        \centering
        \begin{tabular}{l|llllll}
        \toprule
\textbf{Prec.}    &   cifar-10 & f-mnist & speech & 1M & 2M & 5M \\
\midrule
\textbf{16-bit} &  \textbf{0.31}     & 0.09          & 0.0054 & \textbf{0.58}  & \textbf{1.16}   & \textbf{2.90}  \\
\textbf{32-bit} &  0.32     & 0.1           & 0.0048 & 0.70   & 1.40  & 3.52  \\
\textbf{64-bit} & 0.34     & \textbf{0.07}          & \textbf{0.0038} & 0.71  & 1.73  & 3.88 \\
\bottomrule
\end{tabular}
       \caption{For \texttt{topk}, 16-bit provable quantization wins for large data}
       \label{table:quant_topk}
    \end{subtable}
\vspace{-0.2in}
\end{table}

\subsection{Provable Quantization}
\label{subsec:prov_quant_exp}

Provable quantization (\S \ref{sec:quantization}) in \system can optimize the operator memory usage while provably preserving correctness (i.e., no accuracy degradation). To demonstrate its effectiveness, we compare the GPU memory consumption of two applicable operators, \texttt{nwr} and \texttt{topk}, with and without provable quantization using the GPU baseline. Multiple real-world and synthetic datasets are used in the comparison (see Table \ref{table:data}), where synthetic datasets' names, such as ``10k'' and ``1M'', denote their sample sizes. We deem the 64-bit floating point as the ground truth, and evaluate the provable quantization results in 32- and 16-bit floating-point.

The results demonstrate that provable quantization always leads to memory savings. Specifically, Fig.~\ref{fig:quantization_evaluation} (a-b) shows \texttt{nwr} with provable quantization on average saves 71.27\% (with 16-bit precision) and 47.59\% (with 32-bit precision) of the full 64-bit precision GPU memory.
Similarly, Fig.~\ref{fig:quantization_evaluation} (c-d) shows that \texttt{topk} with provable quantization saves 73.49\% (with 16-bit precision) and 49.58\% (with 32-bit precision) of the full precision memory. Regarding the runtime comparison, operating in lower precision may also lead to an edge. Table \ref{table:quantization} shows the operator runtime comparison between using provable quantization (in 16-bit and 32-bit precision) and using the full 64-bit precision. 
\rv{It shows that provable quantization in 16-bit precision is faster than the computations in full precision in most cases, especially for large datasets (e.g., the last three columns of Table \ref{table:quantization}).
This empirical finding can be attributed to lower-precision operations typically being faster, and this speed improvement outweighs the overhead of post verification (see \S \ref{subsec:pq_criteria}). For small datasets (the first three columns of Table \ref{table:quantization}), provable quantization does not necessarily improve the run time due to the additional verification and data movement, both of which finish in 0.1 seconds.}

\begin{table}[!t] 
	\centering
	\caption{Runtime breakdown of using provable quantization on \texttt{nwr} operator with a 30,000 sample synthetic dataset. Column 1 shows the runtime for low-precision evaluation, where column 2 and 3 show the runtime for correctness verification and recalculation, respectively. It shows the primary speed-up comes from low-precision evaluation, while the overhead of verification and recalculation is marginal.} 
	\begin{tabular}{l|lll|l}
		\toprule
		\textbf{Prec.} & \textbf{Low Prec.} & \textbf{Verification} & \textbf{Recalculation} & \textbf{Total} \\
		\midrule
		16-bit              & 5.91       & 0.05          & 0.61 & 6.56           \\
		32-bit              & 22       & 0.05         & 0.47           & 22.51  \\
		64-bit              & N/A      & N/A           & N/A           & 27.32 \\
		\bottomrule
	\end{tabular}
	\label{table:time_breakdown} 
\end{table}

\noindent \MyPara{Case study on runtime breakdown}. In addition to comparing the total runtime of an operator with or without provable quantization (\S \ref{subsec:prov_quant_exp}), it is interesting to see the time breakdown of each phase of provable quantization. Specifically, the runtime of provable quantization can be divided into (\textit{i}) operator evaluation in lower precision, (\textit{ii}) result verification and (\textit{iii}) recalculation in the original precision for the ones that fail in the verification. Taking \texttt{nwr} on 30k samples as an example (the last column of Table \ref{table:quant_nwr}), we show the time breakdown of (\textit{i}) low-precision evaluation (\textit{ii}) correctness verification and (\textit{iii}) recalculation in higher precision in Table \ref{table:time_breakdown}. In this case, the performance improvement of provable quantization comes from the reduced evaluation time in lower precision, which outweighs the cost of verification and recalculation. 
\rv{To further demonstrate the necessity of post-verification, we also measure the accuracy variation (e.g., ROC-AUC \cite{aggarwal2015outlier}) by simply running $k$NN detector on fraud, census, and donors datasets in 16-bit precision without post-verification, which leads to $-3.48\%$, $+1.05\%$, $-4.27\%$ accuracy variation. These results show the merit of provable quantization over direct quantization.}
\vspace{-0.1in}

\subsection{Automatic Batching}
\label{subsec:batching_exp}
To evaluate the effectiveness of automatic batching, we compare the runtime of multiple BOs and FOs under (\textit{i}) an Numpy implementation on a CPU \cite{harris2020array}, (\textit{ii}) a direct PyTorch GPU implementation without batching \cite{paszke2019pytorch}, and (\textit{iii}) \system's automatic batching. 

Table \ref{table:batching} compares the three implementations of key operators in OD systems. Clearly, \system with automatic batching achieves the best balance of efficiency and scalability, leading to 7.22$\times$, 17.46$\times$, and 11.49$\times$ speedups compared to a highly optimized NumPy implementation on CPUs. \system can also handle more than 10$\times$ larger datasets where the direct PyTorch implementation faces out of memory (OOM) errors. \system is only marginally slower than PyTorch when the input dataset is small (see the first row of each operator). In this case, batching is not needed, and \system is equivalent to PyTorch; \system is slightly slower due to the overhead of \system deciding whether or not to enable automatic batching.

\begin{table} 
\centering
	\caption{Operator runtime comparison among implementations in NumPy (no batching), PyTorch (no batching) and \system (with automatic batching); the most efficient result is highlighted in bold per row. Automatic batching in \system prevents out-of-memory (OOM) errors yet shows great efficiency, especially on large datasets.} 
	\vspace{-0.1in}
    \begin{tabular}{ll|lll}
    \toprule
    \textbf{Operator} & \textbf{Size} & \textbf{NumPy} & \textbf{PyTorch} & \system \\
    \midrule
    \texttt{topk}              & 10,000,000       & 7.88          & \textbf{1.08} & 1.09           \\
    \texttt{topk}              & 20,000,000      & 15.77         & OOM           & \textbf{2.44}  \\
    \texttt{topk}              & 100,000,000      & OOM           & OOM           & \textbf{10.83} \\
    \midrule
    \texttt{intersect}         & 20,000,000      & 1.99          & \textbf{0.12} & 0.14           \\
    \texttt{intersect}         & 100,000,000      & 11            & \textbf{0.63} & \textbf{0.63}  \\
    \texttt{intersect}         & 200,000,000      & 21.65         & OOM           & \textbf{2.11}  \\
    \midrule
    \texttt{$k$NN}               & 50,000         & 20.15         & \textbf{0.27} & 0.28           \\
    \texttt{$k$NN}               & 200,000        & 194.43        & OOM           & \textbf{19.65} \\
    \texttt{$k$NN}               & 400,000        & 818.32        & OOM           & \textbf{71.22} \\
    \bottomrule
    \end{tabular}
	\label{table:batching} 
\end{table}

\begin{figure}[!ht] 
\centering
\subfloat{%
  \includegraphics[clip,width=\columnwidth]{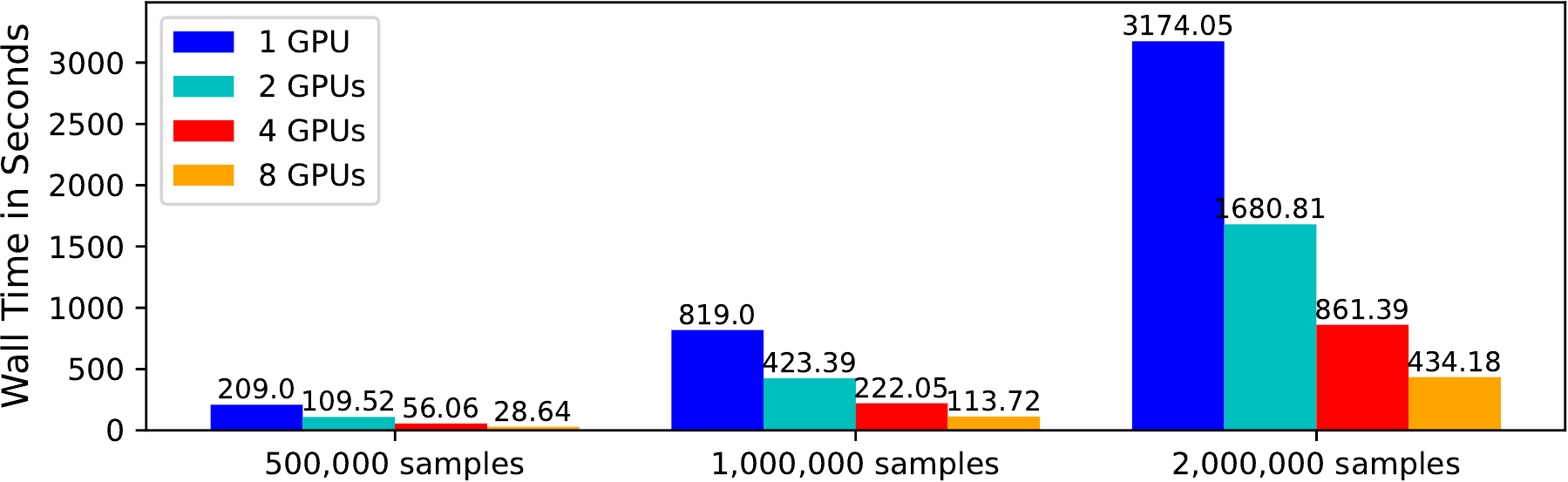}%
}
\vspace{0.1in}

\subfloat{%
  \includegraphics[clip,width=\columnwidth]{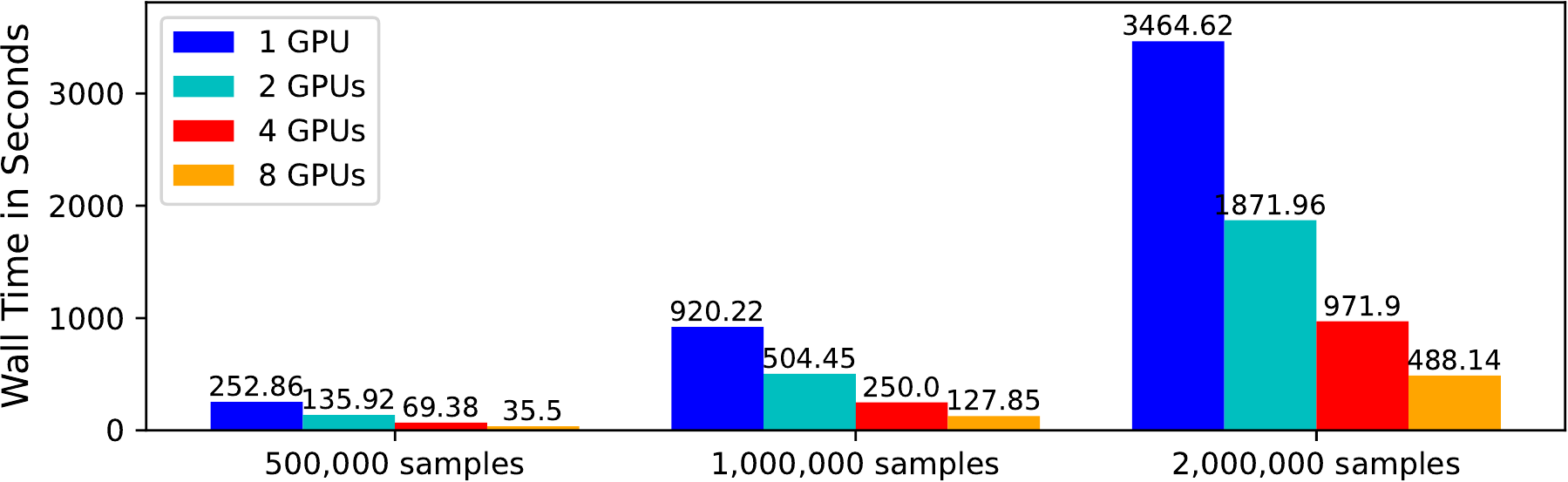}%
}
\vspace{0.1in}

\subfloat{%
  \includegraphics[clip,width=\columnwidth]{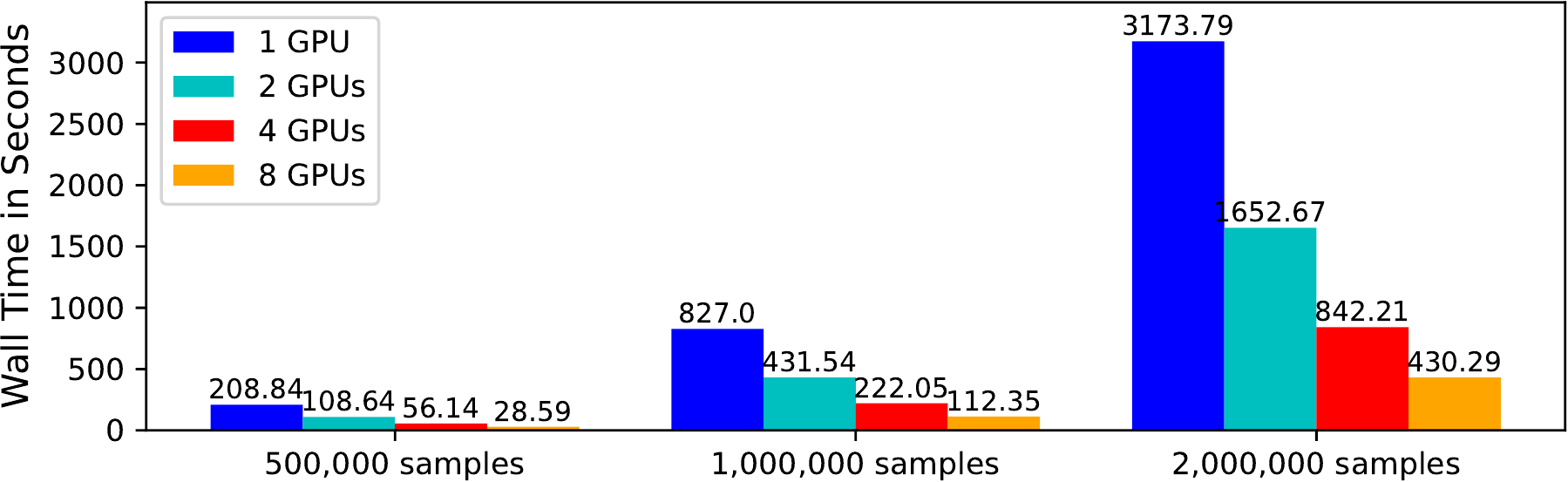}%
}\\

\caption{Runtime comparison of using different numbers of GPUs (\textbf{top}: LOF; \textbf{middle}: ABOD; \textbf{bottom}: $k$NN). \system can efficiently leverage multiple GPUs for faster OD.}
\label{fig:multi_gpu}
\vspace{-0.15in}

\end{figure}

\subsection{Multi-GPU Results}
\label{subsec:multi_gpu}
We now evaluate the scalability of \system on multiple GPUs on a single compute node.
Specifically, we compare the run time of three compute-intensive OD algorithms (i.e., LOF, ABOD, and $k$NN) with 1,2,4, and 8 NVIDIA Tesla V100 GPUs. 

Fig.~\ref{fig:multi_gpu} shows that for three OD algorithms tested, \system can achieve nearly linear speed-up with more GPUs---the GPU efficiency is mostly above 90\%.
For instance, the $k$NN result shows that using 2, 4, and 8 GPUs are $1.91 \times$, $3.73 \times$, and $7.34 \times$ faster than the single-GPU performance. As a comparison, using 2, 4, and 8 GPUs with ABOD and LOF are  $1.85 \times$, $3.63 \times$, $7.14 \times$ faster and $1.91 \times$, $3.70 \times$, $7.27 \times$ faster, respectively. First, there is inevitably a small overhead in multi-processing, causing the speed-up to not exactly be linear. Second, the minor efficiency difference between $k$NN and ABOD is due to most OD operations in the former being executed on multiple GPUs, while the latter involves several sequential steps that have to be run on CPUs. To sum up, \system can leverage multiple GPUs efficiently to process large datasets. 

\vspace{-0.1in}

\section{Limitations and Future Directions}
\label{sec:limitations}
\noindent \rv{\MyPara{Tree-based OD algorithms.}
One limitation of \system is that it does not support tree-based OD algorithms such as isolation forests \cite{DBLP:conf/icdm/LiuTZ08}. Tree-based operations involve random data access \cite{lefohn2006glift}, which is not friendly for GPUs designed for batch operations. Future work can consider converting trees to tensor operations \cite{DBLP:conf/osdi/NakandalaSYKCWI20} for acceleration.}

\noindent \rv{\MyPara{Approximate solutions.}
Our focus in this paper has been on \emph{exact} efficient, scalable implementations of OD algorithms. In particular, we have not considered implementations that intentionally are meant to be approximate, where for instance, one could trade off between accuracy, computation time, and memory usage. Note that even with our provable quantization technique, we ask for the lower-precision computation to yield the correct (exact) output. A future research direction is to extend \system to support approximate solutions for even better scalability and efficiency when reduced accuracy is acceptable. 
For instance, exact nearest neighbor search in 
can be switched to approximate nearest neighbor search \cite{fu2019fast,DBLP:journals/pvldb/WangXY021}.}

\noindent \rv{\MyPara{Heterogeneous GPUs.} Thus far, we have not studied the use of \system with heterogeneous GPUs. In this setting, incorporating a cost model could be helpful in balancing the workload between the different GPUs, accounting for their different characteristics such as varying memory capacities.}

\noindent \rv{\MyPara{Gradient-based operators.} Currently, \system does not support operators that involve solving an optimization problem via gradient descent. Future work may consider incorporating optimization-based operators to support (a small group of) optimization-based OD algorithms, e.g., OCSVM \cite{DBLP:journals/neco/ScholkopfPSSW01}.}

\noindent
\rv{\MyPara{Extension to classification.} It is possible to use \system to build classification models, as the operators in \system are generic and may serve the tasks beyond OD. We provide an example of using \system to construct $k$ nearest neighbor classifier in Appx. \ref{appx:knn}, which also exhibits a significant performance improvement over the CPU implementation in scikit-learn \cite{pedregosa2011scikit}. 
Note that classification tasks often involve optimization (e.g., via gradient descent), which we already pointed out that \system does not yet support. 
Thus, only calculation-based classifiers can be implemented by \system for now.
}
\vspace{-0.2in}

\section{Conclusion}
\label{sec:conclusion}
In this paper, we propose the first comprehensive GPU-based outlier detection system called \system, which is on average \rv{10.9} times faster than the leading system PyOD and is capable of handling larger datasets than existing GPU baselines. The key idea is to decompose complex outlier detection algorithms into a combination of tensor operations for effective GPU acceleration.
Our system enables many large-scale real-world outlier detection applications that could have stringent time constraints. With ease of extensibility, \system can prototype and implement new detection algorithms.

\section*{Acknowledgement}
We would like to thank the anonymous reviewers for their helpful comments. Yue Zhao is partially supported by a Norton Graduate Fellowship. George H.~Chen is supported by NSF CAREER award \#2047981. Zhihao Jia is partially supported by a National Science Foundation award CNS-2147909, and a Tang Endowment.

\bibliographystyle{ACM-Reference-Format}
\bibliography{ref}

\clearpage
\newpage

\appendix

\section*{Supplementary Material for \system}
\textit{Details on system design and experiments.}
\setcounter{table}{0}
\setcounter{figure}{0}
\renewcommand{\thetable}{\Alph{section}\arabic{table}}
\renewcommand{\thefigure}{\Alph{section}\arabic{figure}}

\section{Demo on Using \system to Implement New OD Algorithms from Scratch}
\label{appx:ecod}


\MyPara{Overview of implementing new OD algorithms with \system}. As shown in \S \ref{sec:model_abstraction}, \system takes a modular  approach to implement an OD algorithm. Thus, using \system to implement a new OD algorithm requires
conceptualizing the algorithm as a combination of operators (BTOs and FOs) in \system (see Fig. \ref{fig:operators}). Thanks to the nature of OD applications that do not involve complex optimization, most OD algorithms can be decomposed as a combination of 2 to 3 operators.

\noindent \MyPara{Overview of ECOD}. We provide an example of using \system to implement the latest OD algorithm with empirical cumulative distribution functions (ECOD) \cite{li2022ecod}, which is just published in early 2022. In a nutshell, ECOD estimates the empirical distribution of each feature (i.e., density estimation), where it considers that outliers locate in low-density regions. After that, ECOD aggregates all density estimation results per feature into final outlier scores. 

\begin{figure}[!ht] 
\centering

  \includegraphics[clip,width=0.5\columnwidth]{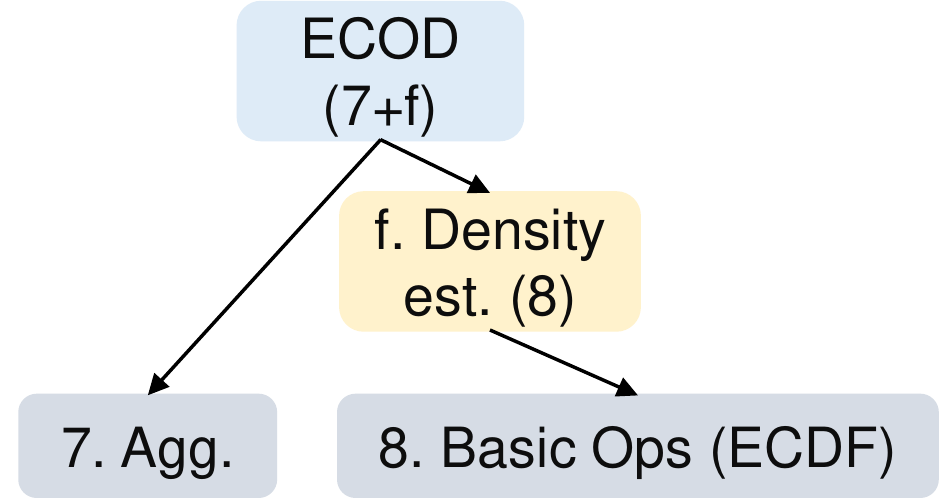}%

\caption{Examples of building latest OD algorithm ECOD with FOs and BTOs conveniently.}
\label{fig:abstraction_examples_appx}
\end{figure}

\noindent \MyPara{Turning ECOD into abstraction.} ECOD mainly involves two operators: (\textit{i}) it estimates the density (``\texttt{f. Density est.}'') of each features by ``\texttt{8. Basic OPs (ECDF)}'' and (\textit{ii}) then aggregates density results across all features via ``\texttt{7. Agg.}''. By having the process in mind, we could sketch the abstraction of ECOD in Fig. \ref{fig:abstraction_examples_appx}.

\noindent \MyPara{Turning abstraction into code}. 
We provide the code breakdown to show how to use \system to implement ECOD, and the full implementation is available for further reference\footnote{ECOD: \url{https://github.com/yzhao062/pytod/blob/main/pytod/models/ecod.py}}. The code snippet (i) shows that we use the basic operator ECDF for density estimation and (ii) shows the aggregation step of density estimation as outlier scores. Putting these core parts together with some skeleton code, we could conveniently include ECOD in \system. It is noted that we still need to glue the operators together, while the use of predefined \system operators helps us to achieve acceleration. 

\color{black}

\begin{center}
    \begin{minipage}{0.9\linewidth}
    \centering
    \begin{lstlisting}[title={(i) code of density estimation in ECOD},captionpos=b]
    from .basic_operators import ecdf_multiple
    
    # density estimation via ECDF
    self.U_l = ecdf_multiple(X, device=self.device)
    self.U_r = ecdf_multiple(-X, device=self.device)
    \end{lstlisting}
    \end{minipage}
\end{center}

\begin{center}
\begin{minipage}{0.9\linewidth}
\centering
\begin{lstlisting}[title={(ii) code of density aggregation in ECOD},captionpos=b]
# take the negative log
self.U_l = -1 * torch.log(self.U_l)
self.U_r = -1 * torch.log(self.U_r)

# aggregate and generate outlier scores
self.O = torch.maximum(self.U_l, self.U_r)
self.decision_scores_ = torch.sum(self.O, dim=1).cpu().numpy() * -1
\end{lstlisting}
\end{minipage}
\end{center}

\section{Example of Using \system to Build Classification Algorithms}
\label{appx:knn}

As discussed in \S \ref{sec:limitations}, \system may be used to construct algorithms beyond OD, e.g., classification and regression. In this section, we demonstrate the use of \system to build a popular classifier called $k$ nearest neighbor classifiers ($k$NN\_CLF)\cite{fix1989discriminatory}. 

In short, $k$NN\_CLF is a lazy classifier and does not involve a  training stage. For a test sample $\mathbf{X}_{\text{test}}$, it calculates the pairwise distance between the test sample with each ``training'' data. It then uses the aggregation (e.g., majority vote or avg.) to decide the predicted classes of each test sample. Thus, it can be decomposed as the combination of (i) calculating pairwise distance by \texttt{cdist} (ii) identifying the top $k$ neighbors by \texttt{topk} and (iii) predicting the class labels by aggregating the labels from the neighbors (i.e., \texttt{Basic OPs}).

\begin{figure}[!ht] 
\centering
  \includegraphics[clip,width=0.5\columnwidth]{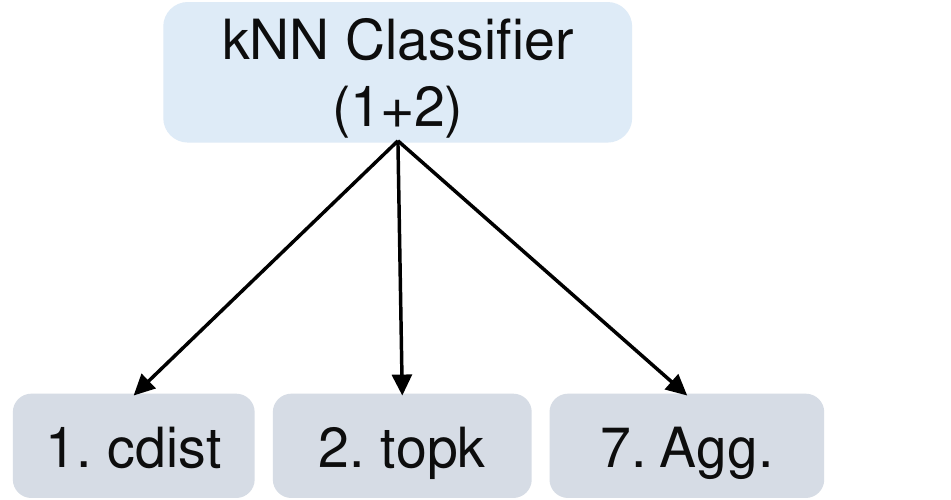}%
\caption{Examples of building $k$NN classifier with \system}
\label{fig:knn_clf}
\end{figure}
\vspace{-0.1in}

Table \ref{table:knn_clf} provides a detailed comparison between CPU-based implementation of $k$NN\_CLF in scikit-learn (sklearn) \cite{pedregosa2011scikit} and \system's implementation. On average, \system's $k$NN\_CLF is 10.1 times faster
than that of sklearn due to the GPU acceleration.

\begin{table}[h] 
\centering
	\caption{Runtime (in seconds) comparison of $k$NN classifier using CPU-based scikit-learn (sklearn) \cite{pedregosa2011scikit} vs. GPU-enabled \system. On average, \system's implementation is 10.1 times faster than that of sklearn.} 

	\footnotesize
	\scalebox{1}{
    \begin{tabular}{lll|ll}
    \toprule
    \textbf{Train Size} & \textbf{Test Size} & \textbf{Number of Features} &\textbf{$k$NN-sklearn} & \textbf{$k$NN-\system} \\
    \midrule
    50,000 & 50,000 & 100 & 41.35 & 4.76\\
    50,000 & 50,000 & 200 & 52.26 & 4.83\\
    100,000 & 50,000 & 100 & 84.84 & 8.09 \\
    100,000 & 50,000 & 200 & 88.13 & 8.36 \\
    \bottomrule
    \end{tabular}
    }
	\label{table:knn_clf} 
\vspace{-0.2in}
\end{table}

\section{Additional Experiment Results}
\label{appx:add_results}

\subsection{End-to-end Results on Synthetic Datasets}
\label{appx:synthetic_results}

\begin{figure}[!h] 
\centering
\subfloat[Synthetic dataset 1 (100,000 samples with 200 features) ]{%
  \includegraphics[clip,width=\columnwidth]{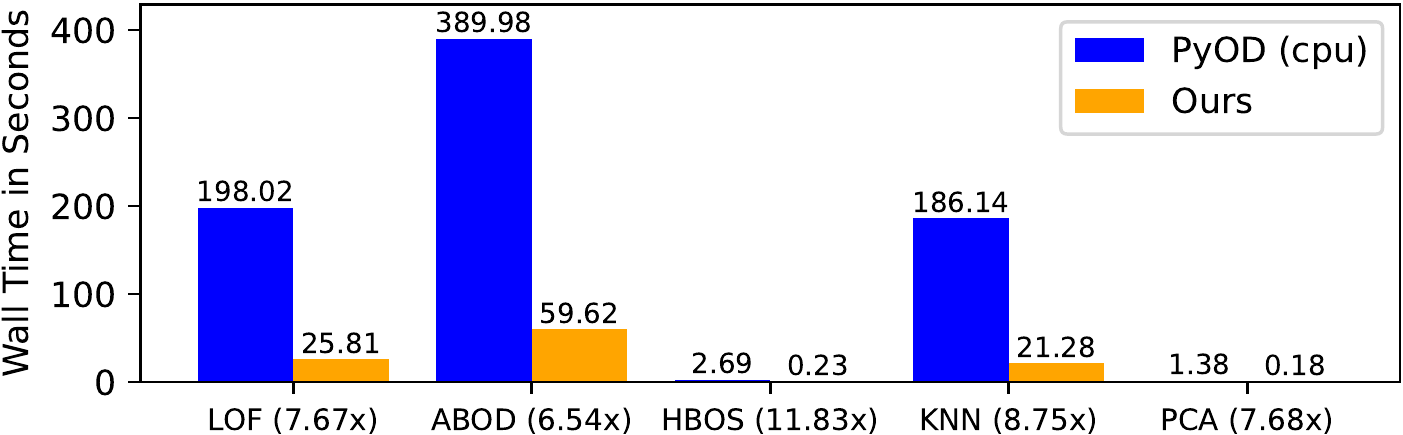}%
}  

\vspace{0.04in}
\subfloat[Synthetic dataset 2 (200,000 samples with 200 features) ]{%
  \includegraphics[clip,width=\columnwidth]{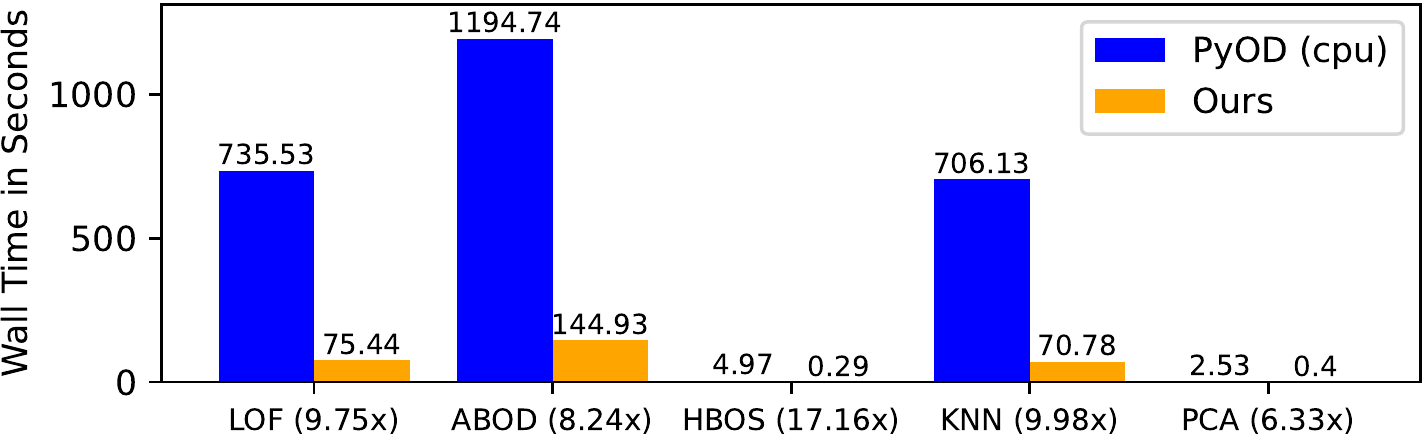}%
} 

\vspace{0.04in}
\subfloat[Synthetic dataset 3 (400,000 samples with 200 features) ]{%
  \includegraphics[clip,width=\columnwidth]{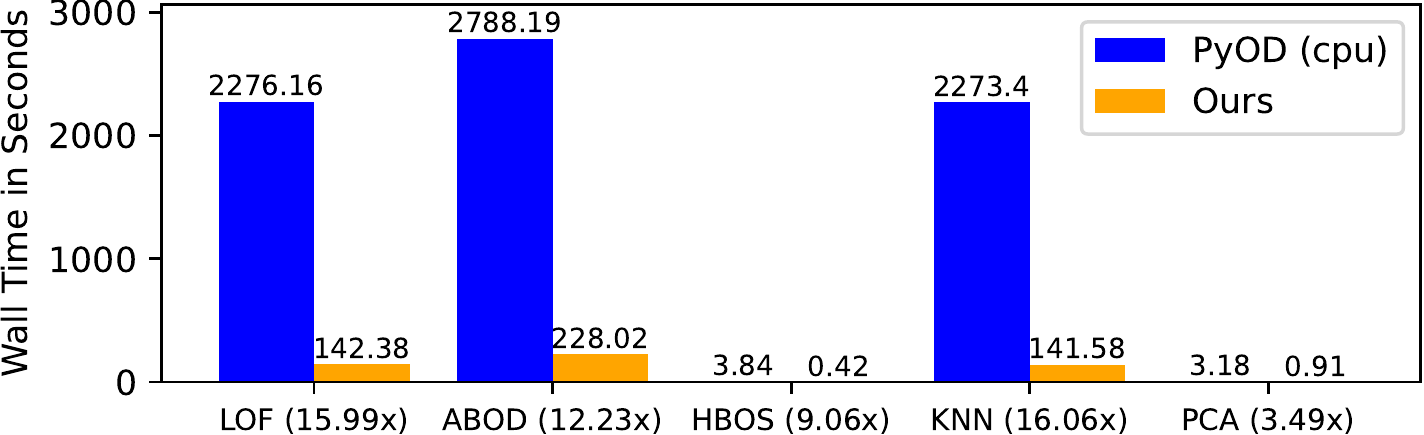}%
} 
\caption{\rv{Runtime comparison between PyOD and \system in seconds on synthetic datasets (see \S \ref{subsec:end_2_end} Fig. \ref{fig:full_system_evaluation} for results on real-world datasets). \system significantly outperforms PyOD in all w/ much smaller runtime, where the speedup factor is shown in parenthesis by each algorithm. On avg., \system is 10.9$\times$ faster than PyOD (up to 38.9$\times$)}.}
\label{fig:full_system_synthetic}
\end{figure}

\rv{Consistent with the results presented in \S \ref{subsec:end_2_end}, \textbf{\system is significantly faster than the leading CPU-based system on large synthetic datasets}. Fig. \ref{fig:full_system_synthetic} shows the results on three synthetic datasets (where Synthetic 1 contains 100,000 samples, Synthetic 2 contains 200,000 samples, and Synthetic 3 contains 400,000 samples (all are with 200 features).
The results show that \system is on average 10.9$\times$ faster than PyOD on the five benchmark algorithms (13.0$\times$, 15.9$\times$, 9.3$\times$, 7.2$\times$, and 8.9$\times$ speed-up on LOF, $k$NN, ABOD, HBOS, and PCA, respectively).}

\subsection{Further Time Analysis}
\label{appx:time_analysis}

\begin{figure}[htp] 
\centering
\subfloat[Synthetic dataset 1 (20,000 samples with 200 features) ]{%
  \includegraphics[clip,width=\columnwidth]{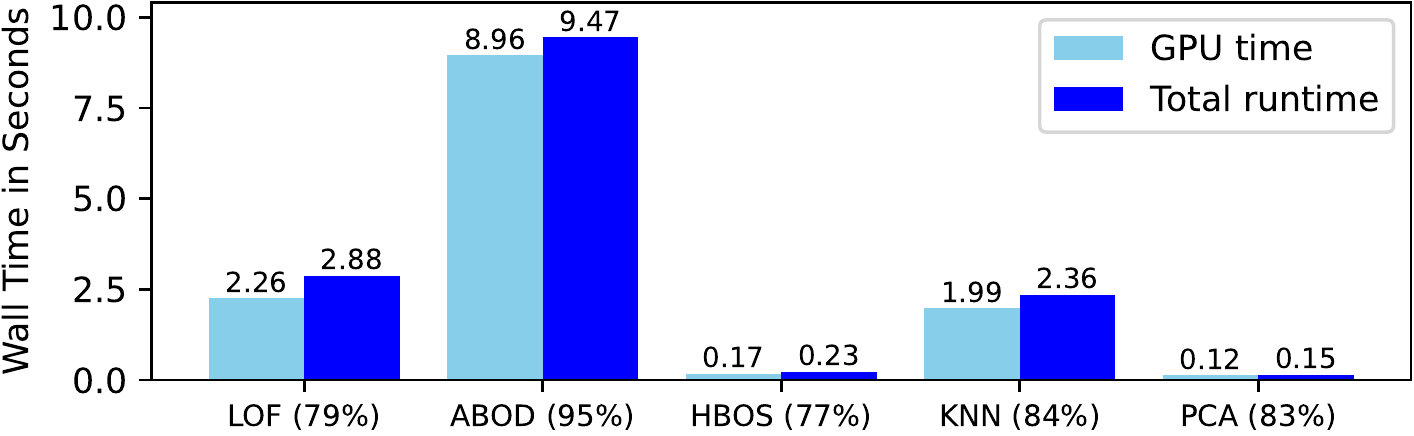}%
}  

\vspace{0.04in}
\subfloat[Synthetic dataset 2 (100,000 samples with 200 features) ]{%
  \includegraphics[clip,width=\columnwidth]{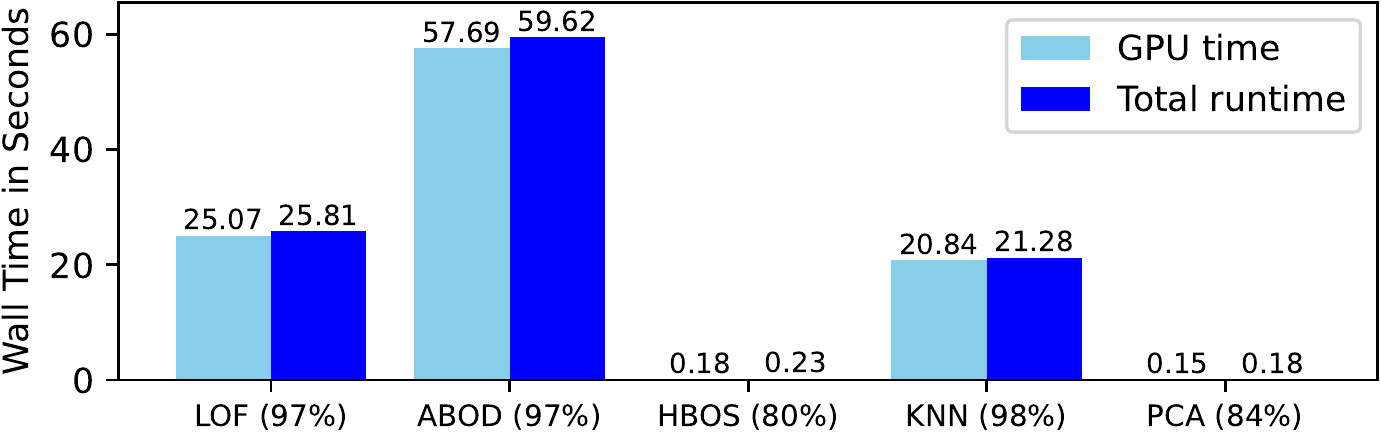}%
} 

\vspace{-0.05in}
\caption{\rv{GPU time and total runtime of \system in seconds on synthetic datasets (where the percentage is shown in parenthesis). \system achieves significant efficiency improvement with high GPU usages (e.g., mostly $>80\%$)}.}
\label{fig:timebreak}
\end{figure}

\rv{Fig. \ref{fig:timebreak} shows the comparison between GPU time and the total runtime, where GPU time consists of a large portion of the total runtime, and verifies that the acceleration of \system mainly comes from the use of GPU(s). Moreover, we also notice that the GPU efficiency increases for larger datasets with higher percentages. This also explains why we observe more acceleration of \system on large datasets in Fig. \ref{fig:full_system_evaluation} and Appx. Fig. \ref{fig:full_system_synthetic}}. 

\subsection{Case Study on Operator Fusion}
\label{appx:fusion_study}
Although automatic batching applies to all BTOs and FOs, we introduce an additional technique called operator fusion in \S \ref{subsec:operator_fusion} to further optimize the execution of selected operators in sequence. For instance, \texttt{$k$NN} batching is achieved by running \texttt{cdist} and \texttt{topk} sequentially, where each uses automatic batching and the output of the former is the input of the latter.

Differently, we could further optimize this simple concatenation execution by fusing \texttt{cdist} and \texttt{topk} together for better efficiency. Fig.~\ref{fig:knn_direct_fusion} compares simple concatenation (subfigure a) and operator fusion (subfigure b) on \texttt{$k$NN}. Specifically, the latter executes the \texttt{topk} BTO on the \texttt{cdist} BTO's individual batches separately rather than running \texttt{topk} on the full distance matrix outputted by \texttt{cdist}. This prevents moving the entire $n \times n$ distance matrix between operators, which often causes OOM.

In Table \ref{table:fusion}, we compare the performance of simple concatenation (i.e., \texttt{cdist} and then \texttt{topk}) and operator fusion regarding both GPU memory and time consumption, where operator fusion shows great performance improvement.

\begin{table}[!ht] 
\centering
	\caption{GPU memory consumption (in Mb) and runtime (in seconds) comparison of simple concatenation (SC) and operator fusion (OF) for \texttt{$k$NN}. Operator fusion brings huge savings in both GPU memory consumption and runtime.} 
	\footnotesize
	\scalebox{1}{
    \begin{tabular}{l|ll|ll}
    \toprule
    \textbf{Sample Size} & \textbf{GPU-SC} & \textbf{GPU-OF} &\textbf{Runtime-SC} & \textbf{Runtime-OF} \\
    \midrule
    50,000 & 5280 &  3060 & 8.47 &  2.80\\
    100,000 &  9010 & 3060 &  33.48 & 9.12 \\

    \bottomrule
    \end{tabular}
    }
	\label{table:fusion} 
\vspace{-0.2in}
\end{table}

\subsection{Ablation Studies on Provable Quantization and Automatic Batching}
\label{appx:ablations}

\rv{In addition to the end-to-end analysis in \S \ref{subsec:end_2_end}, we also provide further ablations on provable quantization (PQ) and automatic batching (AB) in this section.
We demonstrate this experiment with \texttt{$k$-NN} OD algorithm where both techniques apply. Table \ref{table:ablations} shows that automatic batching addresses the issue of out-of-memory on large datasets, while provable quantization reduces GPU memory consumption and runtime. The ablations show that using both techniques leads to the best performance.}

\begin{table}[!ht]
\caption{\rv{Ablations on provable quantization (PQ) and automatic batching (AB) for $k$NN. We set batch size equal to 40,000, and thus AB does not apply to the first row of the table. The best performing combination is highlighted in bold, where using both techniques achieves the best performance.}}
\centering
\label{table:ablations}
    \begin{subtable}[!htb]{0.45\textwidth}
        \centering
        \begin{tabular}{l|llll}
        \toprule
        \textbf{\# Samples} & \textbf{PQ\&AB} & \textbf{AB only} & \textbf{PQ only} & \textbf{None} \\
        \midrule
        40,000              & \textbf{5.34}            & N/A              & 5.34             & 10.88         \\
        80,000              & \textbf{15.12}          & 20.91            & 17.56            & OOM           \\
        160,000             & \textbf{45.37}          & 67.22            & OOM              & OOM          \\
        \bottomrule
        \end{tabular}
       \caption{Runtime in seconds}
       \label{table:ablation_time}
    \end{subtable}
    \hfill
    \begin{subtable}[!ht]{0.45\textwidth}
        \centering
        \begin{tabular}{l|llll}
        \toprule
        \textbf{\# Samples} & \textbf{PQ\&AB} & \textbf{AB only} & \textbf{PQ only} & \textbf{None} \\
        \midrule
40,000              & \textbf{3,040}            & N/A              & \textbf{3040}             & 12,170         \\
80,000              & \textbf{3,040}           & 12,050            & 12,020            & OOM           \\
160,000             & \textbf{3,040}           & 12,050            & OOM              & OOM     \\
        \bottomrule
        \end{tabular}
       \caption{GPU memory consumption in Mb}
       \label{table:ablation_memory}
    \end{subtable}
\end{table}

\color{black}

\section{Open-source System and API Demonstration}
\label{sec:api}
\MyPara{Accessibility}. To facilitate accessibility of \system, we release it under the open BSD 2 license\footnote{Github Repo: \url{https://github.com/yzhao062/pytod}}, which can be easily installed via Python Package Index (PyPI)\footnote{Python Package Index (PyPI: \url{https://pypi.org/project/pytod/}} with name \texttt{pytod}.

\smallskip
\noindent \MyPara{API design and demonstration}. Follow by the mature API design of scikit-learn \cite{sklearn_api} and PyOD \cite{DBLP:journals/jmlr/ZhaoNL19}, all the supported OD algorithms have a unified API design: (\textit{i}) \texttt{fit} processes the input data and calculates necessary statistics for prediction, where the outlier scores are also calculated on the input data (\textit{ii}) \texttt{decision\_function} returns the raw outlier scores of newcoming samples based on the fitted outlier detector in the inductive setting and (\textit{iii}) \texttt{predict} returns the binary labels of newcoming samples based on the fitted outlier detector. We demonstrate these APIs in \system with $k$NN detector below, and other supported OD algorithms have the same APIs.

\begin{center}
\begin{minipage}{0.9\linewidth}
\begin{lstlisting}[title={API demo of invoking $k$NN in \system; other OD algorithms follow the same  API design},captionpos=b, label={lst:label}]
from pytod.utils.data import generate_data
from pytod.utils.data import evaluate_print
from pytod.models.knn import KNN

# Generate sample data
X_train, y_train, X_test, y_test = \
    generate_data(n_train=50000, n_test=10000)

device = validate_device(0)  # get the GPU 0 

# initialize a kNN model in TOD with k=10 and batch=10000
clf = KNN(n_neighbors=k, batch_size=10000, device=device)

# fit the kNN model
clf.fit(X_train)

# get the train labels and outlier scores 
y_train_scores = clf.decision_scores_  # raw outlier scores
y_train_pred = clf.labels_  # binary labels

# evaluate and print the results
evaluate_print(clf_name, y_train, y_train_scores)

# get the prediction labels and outlier scores on test
y_test_scores = clf.decision_function(X_test)  # raw scores
  
\end{lstlisting}
\end{minipage}
\end{center}

\noindent \MyPara{Utility functions.} To facilitate system comparison and evaluation, we also create a set of helper functions including (\textit{i}) \texttt{generate\_data} that can creates synthetic OD datasets by modeling normal samples by Gaussian distribution and outliers by uniform distribution and (\textit{ii}) \texttt{evaluate\_print} to provide OD specific evaluations by the metrics described in \S \ref{subsec:data_base_metrics}.

\label{sec:appendix}

\end{document}